\PassOptionsToPackage{hyphens}{url}
\PassOptionsToPackage{hidelinks}{hyperref}
\documentclass[final]{cas-dc}
\usepackage[numbers,compress]{natbib}

\usepackage{amssymb}
\usepackage{amsmath}
\usepackage{booktabs}
\usepackage{multirow}
\usepackage{siunitx}
\usepackage{subcaption}
\usepackage{algorithm}
\usepackage{algpseudocode}
\usepackage{subfiles}
\usepackage{adjustbox}
\usepackage{placeins}
\usepackage[english]{babel}
\usepackage[final]{microtype}
\usepackage{graphicx}

\usepackage{amsmath,amsfonts,bm}

\def\eqref#1{equation~\ref{#1}}

\def\1{\bm{1}}

\DeclareMathAlphabet{\mathsfit}{\encodingdefault}{\sfdefault}{m}{sl}
\SetMathAlphabet{\mathsfit}{bold}{\encodingdefault}{\sfdefault}{bx}{n}

\graphicspath{{figures/},{./}}

\shortauthors{Nikolay I. Kalmykov et al.}
\shorttitle{T-MLA: A Targeted Multiscale Log--Exponential Attack Framework for Neural Image Compression}

\begin{document}

\title{T-MLA: A Targeted Multiscale Log--Exponential Attack Framework for Neural Image Compression}

\author[1]{Nikolay I. Kalmykov}[orcid=0009-0000-4069-9117]
\ead{nikolay.kalmykov@skoltech.ru}
\author[1]{Razan Dibo}
\ead{razan.dibo@skoltech.ru}
\author[2]{Kaiyu Shen}
\ead{kyshen@std.uestc.edu.cn}
\author[2]{Zhonghan Xu}
\ead{zhonghanxu@stu.uestc.edu.cn}
\author[1]{Anh-Huy Phan}[orcid=0000-0002-5509-7773]\cormark[1]
\ead{a.phan@skoltech.ru}
\cortext[1]{Corresponding author}
\author[2]{Yipeng Liu}
\ead{yipengliu@uestc.edu.cn}
\author[1,3]{Ivan Oseledets}
\ead{i.oseledets@skoltech.ru}

\affiliation[1]{organization={Skolkovo Institute of Science and Technology},
  addressline={Bolshoy Boulevard 30, bld. 1},
  city={Moscow},
  postcode={121205},
  country={Russia}}

\affiliation[2]{organization={University of Electronic Science and Technology of China},
  addressline={No. 2006, Xiyuan Ave, West Hi-Tech Zone},
  city={Chengdu},
  state={Sichuan},
  postcode={611731},
  country={China}}

\affiliation[3]{organization={Artificial Intelligence Research Institute (AIRI)},
  addressline={Presnenskaya Embankment 6, bld. 2},
  city={Moscow},
  postcode={123112},
  country={Russia}}

\begin{abstract}
Neural image compression (NIC) has become the state-of-the-art for rate-distortion performance, yet its security vulnerabilities remain significantly less understood than those of classifiers. Existing adversarial attacks on NICs are often naive adaptations of pixel-space methods, overlooking the unique, structured nature of the compression pipeline. In this work, we propose a more advanced class of vulnerabilities by introducing T-MLA, the first targeted multiscale log--exponential attack framework. We introduce adversarial perturbations in the wavelet domain that concentrate in less perceptually salient coefficients, improving the stealth of the attack. Extensive evaluation across multiple state-of-the-art NIC architectures on standard image compression benchmarks reveals a large drop in reconstruction quality while the perturbations remain visually imperceptible. On standard NIC benchmarks, T-MLA achieves targeted degradation of reconstruction quality while improving perturbation imperceptibility (higher PSNR/VIF of the perturbed inputs) compared to PGD-style baselines at comparable attack success, as summarized in our main results. Our findings reveal a critical security flaw at the core of generative and content delivery pipelines.
\end{abstract}

\begin{keywords}
AI safety \sep adversarial attacks \sep neural image compression \sep frequency-domain analysis
\end{keywords}

\maketitle

\section{Introduction}\label{sec:intro}

\paragraph{Motivation from NIC deployment.}
Classical image codecs, including JPEG~\cite{wallace1991jpeg}, JPEG 2000~\cite{christopoulos2000jpeg2000}, Better Portable Graphics (BPG), and H.266/Versatile Video Coding (VVC)~\cite{bross2021vvc}, rely on fixed transforms and hand-crafted entropy models. However, neural image compression (NIC)~\cite{balle2017endtoend} now matches or exceeds H.266/VVC at 0.15--0.9~bits per pixel (bpp) using standard metrics like Peak Signal-to-Noise Ratio (PSNR) and Visual Information Fidelity (VIF). This progress has led to \textit{JPEG AI's} advancement to Draft International Standard, with real-time 4K decoding on mobile hardware~\cite{jpegai2024dis}. Widespread adoption of NIC across cloud storage, content delivery networks (CDNs), and mobile applications underscores the need to understand its security vulnerabilities~\cite{Wei2021INS}.

\paragraph{Security threats in NIC.}
However, this efficiency creates vulnerabilities~\cite{szegedy2014intriguing}: by back-propagating through the encoder, adversaries can manipulate latents to inflate bitrate~\cite{chen2021robustnic} or corrupt reconstructions~\cite{carlini2017cw}. Recent work~\cite{liu2023malice} showed 56$\times$ bitrate inflation in NIC models, potentially exhausting CDN quotas and mobile data plans, while in safety-critical domains, perturbations can erase medical anomalies~\cite{ma2021understanding} or misalign sensor fusion~\cite{cao2021invisible}. Such adversarial images can also overload on-device decoders~\cite{tabassi2023artificial}.

Throughout this work, we adopt a white-box threat model for NICs: the attacker has full gradient access to a given codec and optimizes codec-specific perturbations. Our primary goal is to provide a controlled robustness assessment of modern open-source NIC architectures, rather than to design a single universally transferable or query-efficient black-box attack across heterogeneous codecs.

\begin{figure}
    \centering
    \includegraphics[width=\columnwidth]{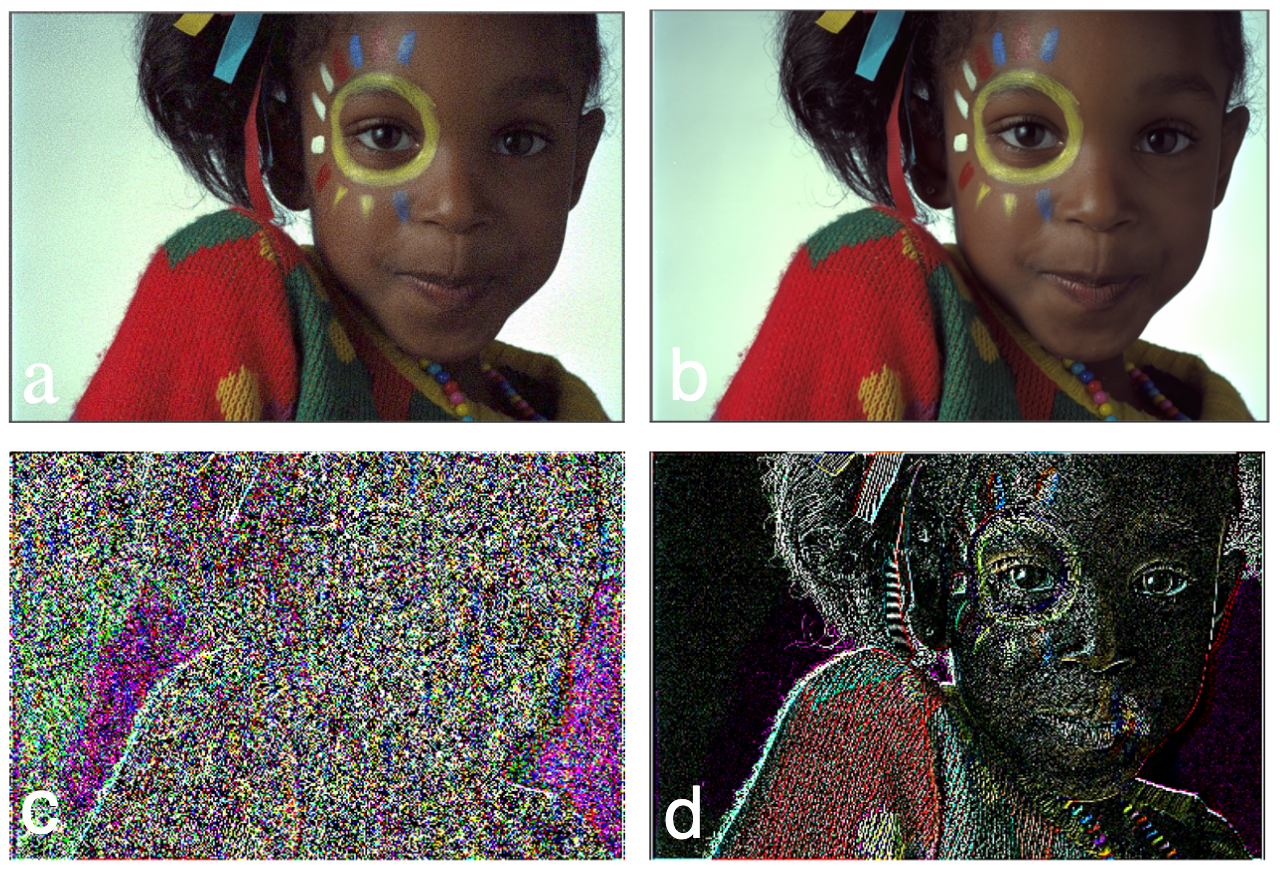}
  \caption{Wavelet-aware adversarial perturbations. (a) PGD-based attack with small-magnitude noise, visually clean yet disruptive after compression. (b) Proposed wavelet-aware attack is also imperceptible but more stealthy. (c) Wavelet coefficients of (a) reveal widespread noise in flat regions. (d) Coefficients of (b) closely resemble the clean input, indicating reduced detectability.}
  \label{fig:wavelet_attack_motivation}
\end{figure}

\paragraph{Limitations of existing attacks.}
Adversarial attacks on neural image compression systems are often implemented through additive perturbations in the spatial domain constrained by the $\ell_\infty$-norm (e.g., Projected Gradient Descent (PGD))~\cite{wang2025fba}, without regard to the frequency structure exploited by modern codecs~\cite{wang2024saliency}. While these attacks may appear imperceptible to humans~\cite{liu2024freqsens}, they often inject widespread high-frequency noise~\cite{Zhao2024INS} that degrades compressed reconstructions.

\autoref{fig:wavelet_attack_motivation}(a) shows an example of such an attack with small-magnitude noise. Visually, the image appears clean, yet it still disrupts the performance of the downstream compression model. To gain further insight into the behavior of the noise, we examine the wavelet coefficients of the attacked image in the high-frequency detail bands, shown in \autoref{fig:wavelet_attack_motivation}(c). These coefficients reveal that the noise is uniformly distributed, even in flat background regions. This suggests that high-frequency wavelet bands can serve as effective indicators of adversarial perturbations.

\paragraph{Our contributions.}
Motivated by this observation, we propose a novel attack method that minimizes perturbation energy not only in the spatial domain but also in the multiscale frequency domain. Our method injects subband-specific noise directly in the wavelet domain, guided by a nonlinear \emph{log--exp mapping} that adapts perturbations according to local coefficient magnitudes. As seen in \autoref{fig:wavelet_attack_motivation}(b, d), our method preserves the appearance of both the image and its wavelet signature, enabling stealthier and more effective attacks.

Our specific contributions are:
\begin{itemize}
    \setlength\itemsep{0pt}
    \setlength\parsep{0pt}
    \setlength\parskip{0pt}
    \setlength\topsep{0pt}
    \item   \textbf{Wavelet-Domain Multiscale Attack:} We propose a novel, frequency-aware attack framework tailored for NIC. Unlike standard additive perturbations, our method employs a multiscale \emph{log--exp mapping} that adapts noise energy to the local magnitude of wavelet coefficients. This design specifically targets the vulnerabilities of entropy models in NIC by hiding perturbations within the natural frequency statistics of the image, thereby achieving high stealth in both spatial and spectral domains (\autoref{sec:wavelet-perturbations}).
    \item \textbf{Comprehensive Evaluation:} We evaluate our attack on texture-diverse images from Kodak, CLIC, and DIV2K datasets (detailed in~\autoref{sec:mean_entropy}). Across Cheng2020-Anchor, Cheng2020-Attention~\cite{cheng2020}, and LIC-TCM~\cite{liu2023tcm} models, T-MLA consistently attains reconstruction PSNR around 24--26~dB under strict stealth constraints of 50--55~dB PSNR at the input (\autoref{fig:psnr_tradeoff}).
    \item \textbf{Ablation Studies and Calibration:} We compare additive, spatial non-linear, and multiscale non-linear attacks across three NIC models (\autoref{tab:model_comparison}), analyze how image characteristics affect attack success (\autoref{sec:mean_entropy}), and report absolute reconstruction PSNR. We further discuss the robustness of NIC models (\autoref{sec:results_and_discussion}).
\end{itemize}

\section{Related Work}
\label{sec:related}

Early adversarial attacks on NICs applied standard computer vision methods: NN codecs~\cite{rippel2017rtic} showed vulnerability to gradient-based attacks, while Generative Adversarial Network (GAN)-based approaches~\cite{mentzer2020hific} introduced new attack surfaces. The work of Chen et al.~\cite{chen2021robustnic} demonstrated that $\ell_\infty$ perturbations could reduce PSNR by 50\% at 0.3~bpp, with Inverse Generalized Divisive Normalization (IGDN) layers amplifying attack effectiveness~\cite{wu2024adversarial}. Liu et al.~\cite{liu2023malice} proposed MALICE, which maximizes entropy in the latent space to increase bitrate by up to 56$\times$ while maintaining visual quality. In contrast to MALICE, which explicitly optimizes for bitrate inflation under relatively mild distortion constraints, our work targets reconstruction degradation at approximately the same bitrate as the original image. In summary, existing NIC attacks predominantly operate in the pixel domain, directly perturbing inputs to the compressor. However, NIC models internally rely on transform-domain representations, which motivates us to next review frequency-domain attacks that manipulate such coefficients more directly.

\noindent\textbf{Frequency-domain adversarial attacks.} Building on classical transform coding, frequency-domain attacks modify Discrete Cosine Transform (DCT) or wavelet coefficients to better exploit the internal structure of compression pipelines. Research in this area~\cite{gu2024freqdomain} has expanded from DCT-based AdvDrop~\cite{duan2021advdrop} to FGL~\cite{wang2024boosting} for mid-frequency targeting, with extensions to backdoor attacks~\cite{tan2023backdoor} and reconstruction distortion~\cite{sui2023reconattack}. AdvWave~\cite{hu2025advwave} exploited multilevel DWT for imperceptible attacks (Structural Similarity Index Measure (SSIM) $\approx$ 0.98), WaveAttack~\cite{li2024waveattack} introduced asymmetric frequency obfuscation for stealthy backdoors, while \cite{wang2025fba} showed frequency methods improve transferability. These approaches demonstrated that targeting specific frequency bands could achieve better imperceptibility than pixel-space perturbations.

Recent benchmarks~\cite{croce2020robustbench} revealed architectural robustness variations, with Cheng2020-Attention showing 15\% better robustness than Cheng2020-Anchor, while JPEG AI~\cite{kovalev2024jpegai} and deeper models exhibit vulnerabilities. Various defenses combine adversarial training~\cite{chen2021robustnic} with robustness analysis~\cite{wu2024adversarial} and gradient obfuscation~\cite{athalye2018obfuscated}, wavelet-based denoising~\cite{wang2021waveletvae} and data-free methods~\cite{mustafa2024wnr}, as well as frequency-domain approaches including diffusion-based defense~\cite{amerehi2025defending}, contrastive learning~\cite{yang2024facl}, and compression-based preprocessing~\cite{jia2019comdefend}. We outline and discuss potential defenses tailored to wavelet-domain attacks in \autoref{alg:defense}.

However, while frequency-domain methods have shown promise in classification tasks, their application to neural image compression remains limited, particularly for achieving compression degradation with visual imperceptibility. For optimization-based baselines, we note that adaptive PGD variants such as APGD~\cite{croce2020apgd} offer stronger step-size control than classical PGD; adapting APGD to the NIC objective is a promising baseline direction, compatible with our frequency-domain design.

\textbf{Scope clarification: wavelet-based attacks in other domains.}
Prior works have explored wavelet-domain adversarial perturbations in tasks unrelated to NIC, such as AdvWave~\cite{hu2025advwave} for style transfer, SITA~\cite{kang2025sita} for classification attacks, and frequency-driven approaches~\cite{luo2022frequency} for evading similarity models. While these methods leverage frequency priors, their goals differ fundamentally from NIC's focus on rate–distortion tradeoffs and entropy model robustness. We do not compare against these approaches, as they fall outside the NIC threat model and exploit vulnerabilities that are not present in compression pipelines. Adapting them would require redesigning their core objectives to target reconstruction quality or bitrate, which is non-trivial and misaligned with their original purpose.

\section{Methodology}\label{sec:methodology}

We propose a wavelet-domain adversarial attack framework that exploits multiscale frequency components of the image. Our approach injects noise directly into the wavelet subbands via nonlinear modulation guided by scale-aware constraints. The goal is to maximize post-compression distortion while maintaining perceptual fidelity (all notations and definitions are provided in~\autoref{sec:appendix}).

\subsection{Problem Setup}

We assume a white-box setting where the attacker has full access to NIC model \( f(\cdot;\theta) \)~\cite{balle2018hyperprior}. Let \( \mathbf{x} \in [0,1]^{I \times J \times C} \) denote the input image and \( \hat{\mathbf{x}} = f(\mathbf{x};\theta) \) the reconstructed output. The objective is to find an adversarial image \( \mathbf{x}_{\mathrm{adv}} \) such that

\begin{equation}
  \max_{\mathbf{x}_{\mathrm{adv}}}\; \mathcal{L}\bigl(f(\mathbf{x}_{\mathrm{adv}}), \mathbf{x}\bigr)
  \quad\text{s.t.}\quad
  \|\mathbf{x}_{\mathrm{adv}} - \mathbf{x}\|_p \le \delta,
  \label{eq:adv_objective}
\end{equation}

\noindent
where \( \mathcal{L} \) denotes a distortion loss, which can be the negative PSNR, and \( \delta \) controls the perturbation magnitude. We deliberately do not include bitrate (BPP) in the objective: all rate values reported later are measured after optimization and serve only as descriptive diagnostics. In this setting T-MLA is optimized per codec, not as a single universal perturbation shared across architectures, which aligns with our goal of white-box robustness assessment for specific NIC models. Classic spatial-domain methods, e.g., Fast Gradient Sign Method (FGSM)~\cite{goodfellow2015explaining}, PGD~\cite{madry2018towards}, and Carlini \& Wagner~\cite{carlini2017cw}, optimize additive noise \( \mathbf{n} \) in the spatial domain

\begin{equation}
  \label{eq:adv_add}
  \mathbf{x}_{\mathrm{adv}} = \mathbf{x} + \mathbf{n},
  \quad \|\mathbf{n}\|_p \le \delta.
\end{equation}

Here, the perturbation \( \mathbf{n} \) is computed via iterative optimization to maximize the loss \( \mathcal{L} \) as specified in~\eqref{eq:adv_objective}. However, these spatial-domain additive perturbations are agnostic to the inherent frequency structure of natural images and often lead to visually noticeable artifacts. They do not exploit the multiscale or directional characteristics of image content, which limits their stealth and adaptability, particularly in systems like NIC, which are sensitive to changes in local image statistics.

\subsection{Wavelet Decomposition}

The \emph{Discrete Wavelet Transform (DWT)} decomposes an image into spatially localized frequency components, providing a multiscale representation. Unlike the Fourier transform's global frequency analysis, DWT offers joint frequency-spatial resolution, making it effective for natural images with localized features like edges and textures.

Formally, given an image \( \mathbf{x} \), a level-\( S \) DWT produces a hierarchical decomposition \(\mathcal{W}(\mathbf{x}) = \left\{ \mathbf{W}_k \right\}_{k=1}^{S+1} \)
where
\[\mathbf{W}_k =
\begin{cases}
\mathbf{H}_k = \{ \mathbf{LH}_k, \mathbf{HL}_k, \mathbf{HH}_k \} & \text{for } k = 1, \dots, S, \\
\mathbf{L}_S & \text{for } k = S+1.
\end{cases}
\]
\noindent
with each \( \mathbf{H}_k \in \mathbb{R}^{I_k \times J_k \times 3C} \) capturing horizontal, vertical, and diagonal detail coefficients at scale \( k \), and \( \mathbf{L}_S \in \mathbb{R}^{I_S \times J_S \times C} \) representing the coarsest approximation.

We use the Haar wavelet (detailed in~\autoref{sec:wavelet}) for its efficiency and simplicity in capturing local intensity changes via averaging and differencing filters, producing sparse multiscale representations (see~\autoref{sec:wavelet_families} for a comparison of wavelet families). Dyadic downsampling yields coarse subbands with fewer but higher-energy coefficients, enabling scale-aware perturbations that remain visually imperceptible while targeting perceptually sensitive regions.

\subsubsection{Intensity and Variance Scaling}

Due to the recursive nature of Haar filtering, the dynamic range of wavelet coefficients expands with scale. Specifically, at scale \( k = 1, \ldots, S\), low-frequency approximation coefficients lie in \(\mathbf{L}_k \in [0, 2^k] \), while high-frequency subbands lie in \( \mathbf{H}_k \in [-2^{k-1}, 2^{k-1}] \).

This behavior introduces a \emph{scaling variance problem}: the magnitude and variance of coefficients naturally increase with scale \( S \), even for clean images. Therefore, comparisons across scales must account for this growth, either through normalization or scale-aware modeling.

\begin{figure*}[t]
  \centering
  \includegraphics[width=\textwidth]{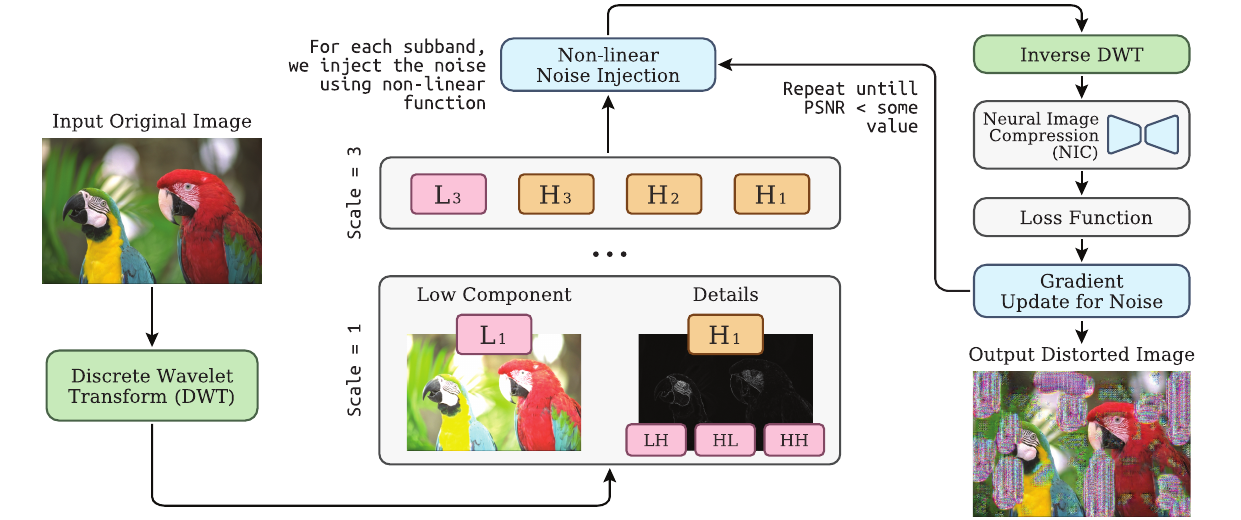}
  \caption{Overview of the proposed T-MLA attack pipeline. The input image after DWT splits into multiple scales, where each scale contains a low-frequency approximation \(\mathbf{L}_S\) and detail components. For visualization, we show the combined magnitude of detail coefficients \(|\mathbf{H}_S| = |\mathbf{LH}_S| + |\mathbf{HL}_S| + |\mathbf{HH}_S|\). The attack iteratively injects nonlinear noise into all subbands, followed by inverse DWT and neural compression to optimize the perturbations that maximize distortion after compression while maintaining visual quality between the original and adversarial images.}
  \label{fig:dwt_explanation}
\end{figure*}

\subsubsection{Implications for Adversarial Detection}

Wavelet detail coefficients \( \mathbf{H}_k \) are sensitive to small, structured perturbations. In natural images, smooth regions yield sparse, low-energy coefficients. Adversarial noise disrupts this by introducing localized high-frequency artifacts, which appear as anomalous magnitudes in the wavelet domain.

\autoref{fig:dwt_explanation} illustrates a 3-scale wavelet decomposition, where we show the combined magnitude of detail coefficients \( |\mathbf{H}_1| \). Such representation better reveals texture patterns compared to individual directional components \(\{\mathbf{LH}_1, \mathbf{HL}_1, \mathbf{HH}_1\}\). Our attack injects noise into the final approximation component \( \mathbf{L}_3 \) and all detail components \( \mathbf{H}_k \).

Unlike pixel-space methods, our wavelet-based attack directly manipulates frequency content. This allows fine-grained, scale- and orientation-aware perturbations that degrade compression quality while remaining visually stealthy, which is key for bypassing detection~\cite{liang2025frenet} and surviving nonlinear NIC pipelines~\cite{xu2025wddsr}.

\subsection{Perturbations in the Multiscale Wavelet Domain}
\label{sec:wavelet-perturbations}

A common class of image perturbations, including many adversarial attacks, is based on \emph{additive noise}. When applied in the wavelet domain, such perturbations take the form of direct modification to wavelet coefficients. Recall that \( \mathcal{W}(\mathbf{x}) = \left\{ \{ \mathbf{H}_k \}_{k=1}^S, \mathbf{L}_S \right\} \) represents the multiscale wavelet decomposition of an image \( \mathbf{x} \). A naive approach perturbs each subband independently 
\[
\widetilde{\mathbf{W}}_k = \mathbf{W}_k + \mathbf{N}_k, \quad \text{for } k = 1, \ldots, S+1
\]
Here, \( \mathbf{N}_k \) denotes additive noise applied to either the low-frequency component \( \mathbf{L}_S \) or the high-frequency detail bands \( \mathbf{H}_k \). The perturbed image is obtained by inverse DWT
\begin{align*}
\mathbf{x}_p &= \operatorname{iDWT}(\widetilde{\mathcal{W}}) = \operatorname{iDWT}(\{\mathbf{W}_k + \mathbf{N}_k\}) \\
&= \operatorname{iDWT}(\mathcal{W}) + \operatorname{iDWT}(\{\mathbf{N}_k\}) = \mathbf{x} + \mathbf{n}
\end{align*}
This shows that additive perturbation in the wavelet domain is functionally equivalent to additive perturbation in the spatial domain, since the DWT is a linear transform. Therefore, this approach fails to exploit the main advantage of wavelet representations, namely, the \emph{scale and direction-based separation} of image features.

\subsubsection{Motivation for Nonlinear Perturbations}

To better align perturbations with perceptually and semantically meaningful structures in the image, we propose a \emph{nonlinear perturbation function} applied in the wavelet domain
\begin{equation}
\widetilde{\mathbf{W}}_k = p(\mathbf{W}_k, \mathbf{N}_k)
= \operatorname{sign}(\mathbf{W}_k)\,\log\!\left(\exp(|\mathbf{W}_k|)+\mathbf{N}_k\right)
\label{eq:logexp_pert}
\end{equation}

This function introduces a nonlinear interaction between the original coefficients \( \mathbf{W}_k \) and the perturbation signal \( \mathbf{N}_k \). It has several desirable properties:
\begin{itemize}
    \setlength\itemsep{0pt}
    \setlength\parsep{0pt}
    \setlength\parskip{0pt}
    \setlength\topsep{0pt}
    \item When \( \mathbf{N}_k = 0 \), we recover the clean coefficients: \( \widetilde{\mathbf{W}}_k = \mathbf{W}_k \).
    \item For large-magnitude coefficients \( |\mathbf{W}_k| \gg 0 \), the perturbation effect is attenuated.
    \item For small-magnitude coefficients \( |\mathbf{W}_k| \approx 0 \), the perturbation has a larger relative effect.
\end{itemize}
Large coefficients in the wavelet domain typically correspond to strong structural features in the image, such as edges and contours. These components are perceptually and semantically important, and perturbing them even slightly can visibly degrade image quality or alter important discriminative cues used by NN. In contrast, small-magnitude coefficients represent weak textures or noise, where perturbations remain imperceptible without changing global image statistics.

Crucially, adding small perturbations to low-magnitude coefficients remains imperceptible while preserving global image statistics. This allows the adversarial noise to hide beneath the ``ground noise floor'' of the image, avoiding both visual and statistical detection.

Thus, the proposed nonlinear perturbation strategy differs from standard additive noise by scaling perturbation strength inversely with the magnitude of the original coefficients. This ensures that attacks are masked within the inherent ``texture'' of the image, exploiting the fact that NIC models often allocate fewer bits to high-frequency, low-energy components. By aligning the attack with these statistical properties, T-MLA achieves a targeted degradation that is structurally stealthy in the multiscale wavelet domain.

This is especially critical because modern NIC models allocate bits via entropy models conditioned on priors that expect smooth decay in high frequencies. Perturbing low-energy detail coefficients disrupts this expectation, effectively ``injecting entropy'' where the model expects redundancy, which forces the decoder to spend bits encoding noise or, when capacity is limited, to drop essential structural information, leading to the observed degradation.

\subsubsection{Approximate Analysis of Perturbation Behavior}

We analyze the perturbation function in \eqref{eq:logexp_pert} applied to a scalar wavelet coefficient \( w \). 
Using a first-order approximation when \( n \ll \exp(|w|) \), we obtain
\begin{align*}
p(w, n) &= \operatorname{sign}(w) \cdot \log\left( \exp(|w|) + n \right)\\
&\approx w + \operatorname{sign}(w) \cdot n \cdot \exp(-|w|)
\end{align*}
This shows that the nonlinear perturbation is approximately equivalent to adding a scaled noise term
\[
n_a = n \cdot \operatorname{sign}(w)  \cdot \exp(-|w|)
\]
The perturbation magnitude decays exponentially with \( |w| \), meaning strong coefficients (large \( |w| \)) are preserved, while weaker, potentially noisy components are more easily perturbed. This allows for a more targeted and perceptually aligned adversarial perturbation strategy in the multiscale domain.

\subsection{Variance-Scaled Noise Budgeting}
\label{sec:variance-scaling}

Adversarial noise is typically constrained to ensure imperceptibility, e.g., by norm bounds such as $\|\mathbf{n}\|_2 \le \delta$ or $\|\mathbf{n}\|_\infty \le \delta$. While global constraints are well-understood in the spatial domain, uniform noise thresholds in the \emph{multiscale wavelet domain} ignore the inherent \emph{scale-variance of coefficients}. Due to recursive downsampling, coefficient magnitudes grow with scale, making uniform constraints either over-perturb low-resolution bands (\( \mathbf{H}_S \), \( \mathbf{L}_S \)) or under-utilize high-resolution ones (\( \mathbf{H}_1 \)). To address this, we propose an \emph{adaptive scaling scheme} for distributing the noise budget across scales.

\begin{algorithm}[t]
  \centering
  \algrenewcommand\alglinenumber[1]{\scriptsize #1:}
  {\small\begin{algorithmic}[1]
    \Require{Image \( \mathbf{x} \), NIC model \( f \), targets \( \mathbf{Q} = (Q_{\text{in}}, Q_{\text{out}}) \), wavelet depth \( S \), noise bound \( \delta \), learning rate \( \eta \), scaling factor \( \alpha \)}
    \State \( \{\mathbf{W}_k\}_{k=1}^{S+1} \gets \operatorname{DWT}^{(S)}(\mathbf{x}) \)
    \State Initialize \( \{\mathbf{N}_k\} \sim \mathcal{N}(0, \delta_k) \), where \( \delta_k = \alpha^{S - k} \cdot \delta \)
    \While{ a stopping condition is not met}
      \For{$k = 1$ \textbf{to} $S+1$}
        \State \( \mathbf{N}_k \gets \operatorname{clip}(\mathbf{N}_k, -\delta_k, \delta_k) \)
        \State \( \widetilde{\mathbf{W}}_k \gets \operatorname{sign}(\mathbf{W}_k) \cdot \log\left(\exp(|\mathbf{W}_k|) + \mathbf{N}_k\right) \)
      \EndFor
      \State \( \mathbf{x}_{\mathrm{adv}} \gets \operatorname{iDWT}^{(S)}(\{\widetilde{\mathbf{W}}_k\}) \)
      \State \( \hat{\mathbf{x}}_{\mathrm{adv}} \gets f(\operatorname{clip}(\mathbf{x}_{\mathrm{adv}}, 0, 1)) \)
      \For{$k = 1$ \textbf{to} $S+1$}
        \State \( \mathbf{N}_k \gets \mathbf{N}_k - \eta \cdot \nabla_{\mathbf{N}_k} \mathcal{L} \)
      \EndFor
    \EndWhile
    \Ensure{\( \mathbf{x}_{\mathrm{adv}}, \hat{\mathbf{x}}_{\mathrm{adv}}, \{\mathbf{N}_k\} \)}
  \end{algorithmic}}
  \caption{Targeted Multiscale Log--Exp Attack (T-MLA)}\label{alg:tmla}
\end{algorithm}

\subsection{Adaptive Noise Bounds per Subband}

To account for scale-dependent variation in wavelet coefficient magnitudes, we apply an \emph{adaptive} \( \ell_\infty \) constraint on the noise in each subband:
\[
\|\mathbf{N}_k\|_\infty \leq \delta_k, \quad \text{where} \quad \delta_k = \alpha^{S - k} \cdot \delta.
\]
Here, \( \delta \) is the global noise budget, and \( \alpha \) controls how the
bound scales. This scaling strategy applies to \emph{all} subbands (including the approximation band \(\mathbf{L}_S\)), ensuring a unified optimization process that generalizes across different compression architectures without requiring model-specific subband selection. While theory suggests \( \alpha = 2 \) based on coefficient variance doubling at finer scales, empirical analysis on DIV2K and CLIC datasets yields \( \alpha \approx 1.9 \). We use \( \alpha = 1.8 \) to balance imperceptibility and attack effectiveness (detailed analysis is provided in~\autoref{sec:empirical-alpha}).

\subsection{Targeted Multiscale Log--Exp Attack (T-MLA)}

We define a composite loss for learning the adversarial noise in the wavelet domain 
\[
\begin{aligned}
\mathcal{L} &= \left|\text{PSNR}(\hat{\mathbf{x}}_{\mathrm{adv}}, \mathbf{x}) - Q_{\text{out}}\right| \\
&\quad + \left|\text{PSNR}(\mathbf{x}_{\mathrm{adv}}, \mathbf{x}) - Q_{\text{in}}\right| 
+ \lambda \sum_k \|\mathbf{N}_k\|_1
\end{aligned}
\]
where \( Q_{\text{out}} \) controls compression attack strength (distort post-codec
output), \( Q_{\text{in}} \) controls visual deviation, and \( \ell_1 \)-regularization encourages sparse perturbations. The noise tensors \( \mathbf{N}_k \) are optimized jointly using Adam optimizer, with clipped noise and inverse DWT (iDWT) reconstructing \( \mathbf{x}_{\mathrm{adv}} \) before passing through NIC to obtain \( \hat{\mathbf{x}}_{\mathrm{adv}} \).

In this work, the term ``targeted'' specifically refers to driving the codec output towards a desired reconstruction PSNR level \(Q_{\text{out}}\) (while maintaining a desired input PSNR level \(Q_{\text{in}}\)), rather than targeting a particular semantic class, spatial region, or high-level visual attribute.

Mathematically, this formulation differs from conventional distortion/perceptual losses (e.g., plain \(\ell_2\), SSIM-, or LPIPS-based objectives) in two ways. First, the PSNR terms act as soft \emph{targets} that pin the optimization to a specific operating point \((Q_{\text{in}},Q_{\text{out}})\), rather than merely encouraging ``as much distortion as possible'' under a global norm bound. Second, in combination with the log--exp perturbation in~\eqref{eq:logexp_pert} and its approximation in the analysis above, the effective additive noise is reweighted by a factor \(\exp(-|w|)\) in the wavelet domain, yielding a content- and scale-aware loss landscape that penalizes deviations on large-magnitude, structure-carrying coefficients much more strongly than on low-energy ones; standard \(\ell_2\)/SSIM/LPIPS losses applied in pixel space lack this built-in, coefficient-dependent weighting across scales.

Our multiscale adversarial attack is summarized in \autoref{alg:tmla}. At each iteration, noise is added adaptively across wavelet subbands under scale-aware constraints. The perturbations are applied in the wavelet domain using a log-exp formulation, then transformed back to pixel space via inverse DWT. The resulting adversarial image is passed through the NIC model, and the perturbation is updated to maximize a compression-targeted loss while maintaining imperceptibility.

\section{Experimental Setup}\label{sec:experimental_setup}

\begin{figure*}[!t]
  \centering
  \includegraphics[width=0.75\textwidth]{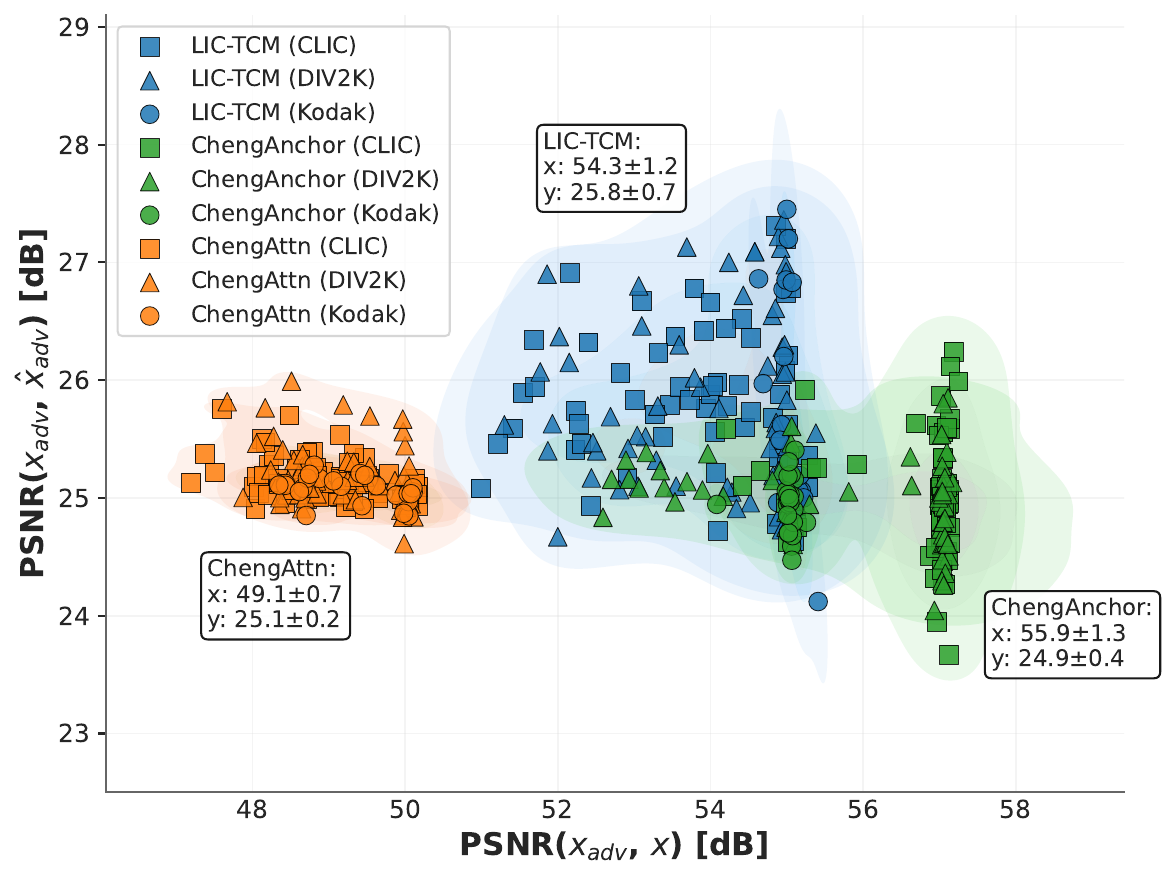}
  \caption{Stealth vs.\ degradation tradeoff for T-MLA across NIC models and datasets. Each point represents a model-image pair and the corresponding attack performance. The axes are \( \mathrm{PSNR}(x_{\mathrm{adv}}, x) \uparrow \) vs.\ \( \mathrm{PSNR}(x_{\mathrm{adv}}, f(x_{\mathrm{adv}})) \downarrow \) (targets: 50/55~dB and 25~dB).
  }
  \label{fig:psnr_tradeoff}
\end{figure*}

\subsection{Datasets and Target Models}
We evaluate on three standard image compression datasets, using representative images from each.
\emph{Kodak}~\cite{franzen1999kodak} contains 24 classical $768 \times 512$ PNGs with diverse textures. 
\emph{CLIC-pro}~\cite{toderici2020clic} has 250 high-resolution professional photos, from which we select 63 images matched to Kodak's local entropy distribution, then crop to $768 \times 512$ (or $512 \times 768$). 
\emph{DIV2K}~\cite{agustsson2017ntire} includes 800 images at $2{,}048 \times 1{,}356$; we select 65 entropy-matched samples. 
Selection ensures comparable complexity across datasets (see \autoref{fig:entropy-dist}).

We test on three NIC models: 
\emph{Cheng2020-Anchor} and \emph{Cheng2020-Attention}~\cite{cheng2020}, which use hyperprior entropy modeling (with and without attention), and 
\emph{LIC-TCM}, a recent transformer–CNN hybrid~\cite{liu2023tcm}. 

\subsection{Baseline and Evaluation Metrics}
We compare our method (\textbf{T-MLA}) against: \emph{Pixel-Additive (PGD)}: a standard pixel-space PGD attack~\cite{chen2021robustnic}; and \emph{Pixel-LogExp}: our method with $S{=}0$ and no wavelet 
stages, to isolate the multiscale effect.

We report the following metrics (\autoref{tab:model_comparison}). Stealth measures perturbation imperceptibility using PSNR and VIF between $\mathbf{x}$ and $\mathbf{x}_{\mathrm{adv}}$ ($\uparrow$ better). Attack success measures reconstruction degradation using PSNR, SSIM, and VIF between $\mathbf{x}$ and $\hat{\mathbf{x}}_{\mathrm{adv}}$ ($\downarrow$ better). We also report bitrate (BPP).

\subsection{Implementation Details}
PyTorch 2.7.1 (CUDA 12.6) with CompressAI library for NIC and `pytorch-wavelets` package; hardware: A100-SXM4-40GB ($\times$2) and RTX 4090 workstation. Unless otherwise stated, all experiments use a fixed random seed of 42. For \autoref{alg:tmla} we use Haar wavelets ($S=3$ scales), Adam optimizer ($\eta=1\text{e-}2$), and perturbation budget $\delta=0.03$ with scale-adaptive constraints $\|\mathbf{N}_k\|_\infty \leq \delta_k$. We set $Q_{\text{in}}=55\pm2$ dB for Cheng2020-Anchor and LIC-TCM, $Q_{\text{in}}=50\pm2$ dB for Cheng2020-Attention, and $Q_{\text{out}}=25\pm2$ dB for all models. On an A100 GPU, a full T-MLA run for a $768{\times}512$ Kodak image takes about 20 minutes for Cheng2020-Anchor (roughly 2.5\,s per 100 iterations), with comparable runtime for Cheng2020-Attention and about $5\times$ slower iterations for LIC-TCM; peak memory usage is approximately 2.0\,GB (Anchor), 2.5\,GB (Attention), and 11.0\,GB (LIC-TCM).

\begin{table*}[t]
  \centering
  \caption{Attack performance on NIC models (avg. over Kodak, CLIC, DIV2K). Best stealth scores in \textbf{bold}. Stealth: higher PSNR/VIF is better. Attack success: lower PSNR/SSIM/VIF indicate stronger attack. BPP is reported descriptively; no rate objective is optimized.}
  \label{tab:model_comparison}
  {\small\setlength{\tabcolsep}{16pt}
  \begin{adjustbox}{max width=\textwidth}
  \begin{tabular}{@{}lccccccc@{}}
    \toprule
    \textbf{Method} & \textbf{Model} & \multicolumn{2}{c}{\textbf{Stealth} $\uparrow$} & \multicolumn{4}{c}{\textbf{Attack Success} $\downarrow$}  \\
    \cmidrule(lr){3-4} \cmidrule(lr){5-8}
    & & \textbf{PSNR} & \textbf{VIF} & \textbf{PSNR} & \textbf{SSIM} & \textbf{VIF} & \textbf{BPP}\\
    \midrule
    \multirow{3}{*}{PGD}
    & Cheng2020-Anchor    & 40.15 ± 0.02& 0.973 ± 0.018 & 8.89 ± 2.73 & 0.691 ± 0.103 & 0.081 ± 0.137 & 0.93 ± 0.37 \\
    & Cheng2020-Attention & 30.67 ± 0.04& 0.815 ± 0.076 & 6.39 ± 3.24 & 0.583 ± 0.081 & 0.065 ± 0.181 & 1.52 ± 0.21 \\
    & LIC-TCM             & 34.37 ± 0.07& 0.918 ± 0.044 & 11.00 ± 3.17 & 0.717 ± 0.073 & 0.127 ± 0.165 & 6.73 ± 1.20 \\
    \midrule
    \multirow{3}{*}{LogExp}
    & Cheng2020-Anchor    & 43.42 ± 0.33 & 0.984 ± 0.004 & 21.32 ± 1.52 & 0.910 ± 0.090 & 0.488 ± 0.117 & 0.87 ± 0.35 \\
    & Cheng2020-Attention & 37.04 ± 0.09 & 0.968 ± 0.006 & 32.07 ± 0.78 & 0.944 ± 0.056 & 0.929 ± 0.169 & 0.93 ± 0.33 \\
    & LIC-TCM             & 36.69 ± 0.75 & 0.946 ± 0.005 & 14.74 ± 1.42 & 0.825 ± 0.073 & 0.223 ± 0.106 & 4.40 ± 0.70 \\
    \midrule
    \multirow{3}{*}{T-MLA}
    & Cheng2020-Anchor    & \textbf{55.82 ± 0.87} & \textbf{0.999 ± 0.001} & 25.03 ± 0.33 & 0.969 ± 0.010 & 0.728 ± 0.140 & 0.85 ± 0.37 \\
    & Cheng2020-Attention & \textbf{49.14 ± 0.70} & \textbf{0.996 ± 0.003} & 25.20 ± 0.20 & 0.970 ± 0.009 & 0.839 ± 0.107 & 0.86 ± 0.37 \\
    & LIC-TCM             & \textbf{54.31 ± 0.83} & \textbf{0.999 ± 0.001} & 25.71 ± 0.73 & 0.963 ± 0.011 & 0.735 ± 0.130 & 1.35 ± 0.48 \\
    \bottomrule
  \end{tabular}
  \end{adjustbox}}
\end{table*}

\section{Results and Discussion}\label{sec:results_and_discussion}

\subsection{Quantitative Analysis}

\autoref{tab:model_comparison} summarizes performance across three NIC models, with results for each method-model combination averaged over the three datasets (Kodak, CLIC, DIV2K). T-MLA markedly improves stealth PSNR over pixel-space PGD while maintaining attack strength: \emph{Cheng2020-Anchor} 55.82 vs.\ 40.15~dB ($\uparrow$15.67), \emph{Cheng2020-Attention} 49.14 vs.\ 30.67~dB ($\uparrow$18.47), \emph{LIC-TCM} 54.31 vs.\ 34.37~dB ($\uparrow$19.94).
For a per-codec statistical comparison on the Kodak dataset, including mean $\pm$ standard deviation and significance tests, see the bar plots in~\autoref{fig:tmla_vs_pgd_kodak}. For a stronger pixel-space baseline on Cheng2020-Anchor, we additionally compare against APGD~\cite{croce2020apgd} on Kodak; summary statistics are provided in~\autoref{tab:apgd_anchor_kodak_stats}. On this setting, APGD slightly improves stealth over PGD (input PSNR $42.10$ vs.\ $40.15$~dB) but largely sacrifices attack strength (output PSNR increases from $8.89$ to $25.18$~dB), whereas T-MLA achieves much higher imperceptibility (PSNR $55.82$~dB, VIF $0.999$) while keeping the output PSNR around $25$~dB, yielding a strictly better stealth--success trade-off.

\subsubsection{Stealth–Success Tradeoff Visualization}

T-MLA satisfies both stealth and attack constraints across datasets. \autoref{fig:psnr_tradeoff} provides full distributions over all images. Kodak, CLIC and DIV2K samples tightly cluster around the stealth target (\( Q_\text{in} = 50/55 \)~dB) within a ±2~dB tolerance and the attack strength (\( Q_\text{out} = 25 \)~dB) is consistently achieved across datasets. Slightly higher degradation PSNR (i.e., worse attack) is acceptable if stealth is preserved.

Despite similar degradation (\(\sim25\text{--}26\)~dB), we see that \emph{Cheng2020-Attention} is most robust (stealth PSNR \(\sim49\)~dB); \emph{Cheng2020-Anchor} achieves the same degradation with smaller visible perturbation (stealth PSNR \(\sim56\)~dB); \emph{LIC-TCM} is comparatively vulnerable (stealth PSNR \(\sim54\)~dB), indicating susceptibility of its hybrid architecture.

\subsubsection{Robustness vs.\ Local Entropy}
\begin{figure*}[!t]
    \centering
    \begin{subfigure}[t]{0.49\textwidth}
        \centering
        \includegraphics[width=\linewidth]{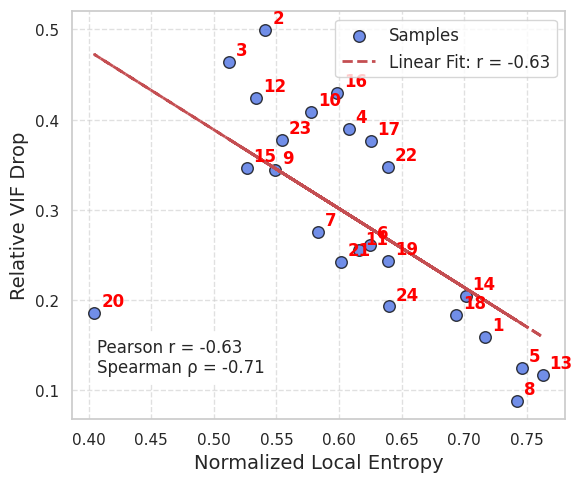}
        \caption{
        Relative VIF drop vs.\ entropy for Kodak (Cheng2020-Anchor). 
        Lower entropy $\rightarrow$ larger perceptual loss.
        }
        \label{fig:entropy_vifdrop_anchor}
    \end{subfigure}
    \hfill
    \begin{subfigure}[t]{0.49\textwidth}
        \centering
        \includegraphics[width=\linewidth]{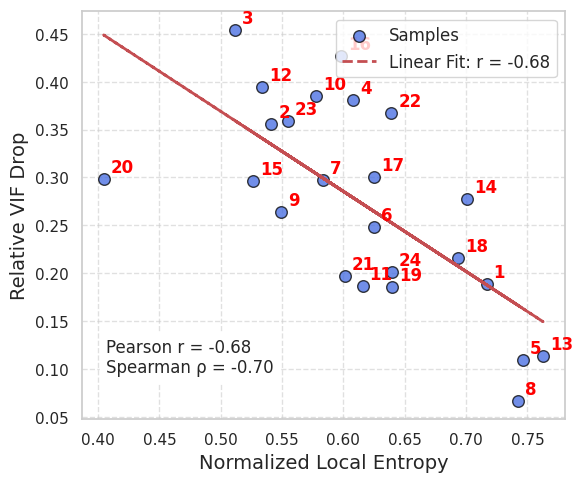}
        \caption{
        Relative VIF drop vs.\ entropy for Kodak (TCM). 
        Similar trend: high-entropy images are more robust.
        }
        \label{fig:entropy_vifdrop_tcm}
    \end{subfigure}
    \caption{
    Entropy-dependent robustness (relative VIF drop) for two models.
    }
    \label{fig:combined_figures}
\end{figure*}

We further analyze the relationship between image characteristics and adversarial robustness by comparing the relative VIF drop (difference between stealth VIF and attack VIF) against normalized local entropy, focusing on Kodak samples. \autoref{fig:combined_figures}(a,b) shows this correlation for two NIC models and for both models, we observe a strong negative correlation: \emph{Cheng2020-Anchor} (Pearson \( r = -0.63 \), Spearman \( \rho = -0.71 \)) and \emph{LIC-TCM} (Pearson \( r = -0.68 \), Spearman \( \rho = -0.70 \)). This indicates that \emph{images with lower local entropy}, i.e., smoother or more homogeneous textures, exhibit a \emph{larger relative drop in VIF} under adversarial attack. In contrast, \emph{high-entropy images} (e.g., those with texture and detail) are slightly more robust in preserving fine perceptual structure, possibly due to masking effects.

To further generalize this relationship, we analyzed the relative VIF drop as a function of mean entropy across entire datasets, including CLIC, DIV2K, and Kodak (see~\autoref{sec:appendix}). The results reveal a consistent negative correlation between entropy and attack effectiveness: Pearson correlations of \( r = -0.241 \) (\( p < 0.001 \)) for CLIC and DIV2K, and \( r = -0.579 \) (\( p < 0.001 \)) for Kodak.

These findings suggest that entropy-aware metrics may help predict where adversarial perturbations will have the most visual impact, and could inform entropy-adaptive defenses in future NIC models.

\subsubsection{Optimization Convergence and Efficiency}

We next analyze the convergence behavior and computational cost of T-MLA. A detailed convergence analysis, including $Q_{\text{in}}$ and $Q_{\text{out}}$ versus iterations for a representative case (kodak23, Cheng2020-Anchor), is provided in~\autoref{sec:convergence_analysis}. The plots show that the input PSNR is increased in stages towards the 55~dB target while the output PSNR remains concentrated around 25~dB, indicating stable optimization under the joint stealth--degradation objective. Precise runtimes and GPU memory usage for all three NIC models are summarized in~\autoref{sec:experimental_setup}.

\subsubsection{Perceptual Metrics Analysis}
\label{sec:perceptual_metrics}
To assess imperceptibility beyond Mean Squared Error (MSE)-based metrics, we additionally evaluated our method using Learned Perceptual Image Patch Similarity (LPIPS) and Deep Image Structure and Texture Similarity (DISTS) on the Kodak dataset with Cheng2020-Anchor. As shown in~\autoref{fig:imperceptibility}, T-MLA achieves a mean LPIPS of 0.0028$\pm$0.0016, consistently remaining well below the visibility threshold of 0.01 (max 0.0065) and nearly two orders of magnitude lower than PGD (0.545). The massive improvement in PSNR correlates with near-perfect perceptual scores (VIF 0.9995, DISTS 0.0008), confirming the invisibility of our multiscale perturbations.

\subsection{Qualitative Analysis}
\begin{figure*}[t]
    \centering
    \includegraphics[width=\textwidth]{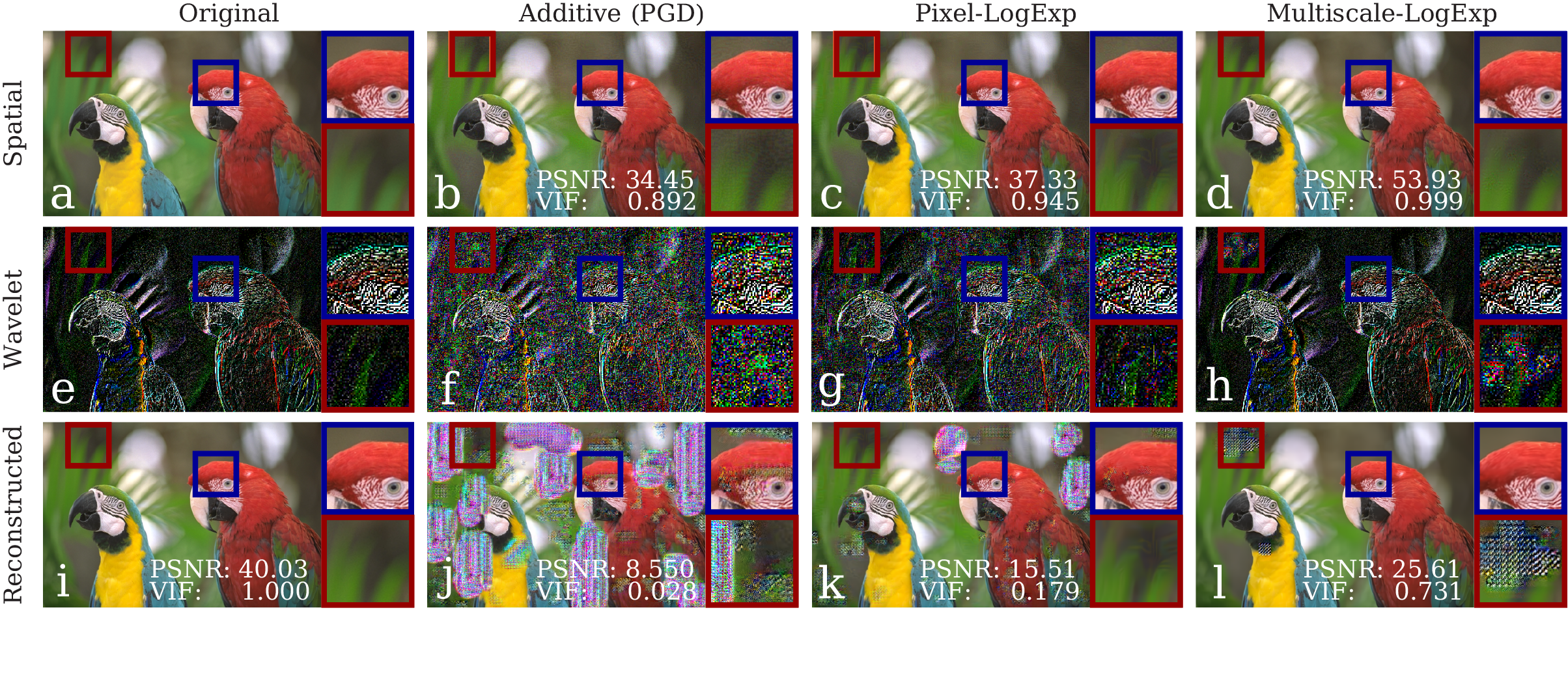}
    \caption{Comprehensive visual comparison of adversarial attacks against LIC-TCM model for Kodak image 23. Top row (a-d): Original image and perturbed images using Additive (PGD), Pixel-LogExp, and T-MLA attacks. Middle row (e-h): Corresponding wavelet-domain representations showing original coefficients and attack-induced perturbations. Bottom row (i-l): Original reconstructed image and reconstructed images after compression under each attack method. The T-MLA approach demonstrates targeted frequency-domain perturbations while achieving severe reconstruction degradation.}
    \label{fig:full_visual_comparison}
  \end{figure*}

\autoref{fig:full_visual_comparison} compares across three domains: spatial, frequency, and reconstructed output. In the spatial domain (top row), all attack methods produce visually imperceptible perturbations, with subtle differences only visible in the magnified insets (blue and red boxes). However, the frequency domain analysis (middle row) reveals distinct signatures in attack characteristics.

The additive attack in \autoref{fig:full_visual_comparison}(f) introduces widespread high-frequency artifacts across all wavelet subbands, while Pixel-LogExp in \autoref{fig:full_visual_comparison}(g) shows reduced but still substantial spectral noise. In contrast, our T-MLA method in \autoref{fig:full_visual_comparison}(h) produces minimal frequency-domain disturbance, a critical advantage for evading spectral anomaly detection systems. This difference is particularly evident in the magnified frequency insets: while baseline methods show obvious perturbation patterns even to the naked eye, T-MLA perturbations are only discernible in the red inset, with the blue inset region appearing nearly identical to the original.

The reconstruction quality (bottom row) demonstrates T-MLA's attack effectiveness despite its frequency stealth. While classical methods achieve stronger visual degradation in the reconstructed output, our method achieves the optimal trade-off between imperceptibility and attack success, which is a crucial requirement for practical adversarial scenarios where detection avoidance is essential.

\subsection{Defense Evaluation}
We explored a wavelet-aware defense mechanism to counter T-MLA, conceptually mirroring the attack by minimizing distortion via gradient descent in the wavelet domain (detailed in~\autoref{sec:defense}). While a full benchmark of defense strategies (e.g., adversarial training, frequency anomaly detection) is beyond our scope, preliminary results (see~\autoref{fig:defended-div-090}) show that this adaptive defense can effectively restore visual quality (PSNR/SSIM) by neutralizing adversarial perturbations in the frequency bands they target. This confirms that understanding wavelet-domain vulnerabilities is key to designing robust NIC systems.

\subsection{Ablation Study}
We analyzed different scale levels ($S=1,2,3$) for wavelet decomposition. While PSNR values remained comparable across scales, higher scales naturally concentrated perturbations in textured regions, improving visual imperceptibility (detailed analysis in~\autoref{sec:wavelet}). Moreover, additional visual comparisons and detailed quantitative results are provided in Appendix. In line with \autoref{fig:scales-a}, $S=2$ often balances stealth and degradation; $S=3$ offers slightly better frequency visual stealth at similar targets, which we adopt by default.

\section{Conclusion}\label{sec:conclusion}

We introduced T-MLA, a targeted multiscale wavelet-domain attack that exploits vulnerabilities in neural image compression (NIC) models. By injecting frequency-specific perturbations, T-MLA achieves substantial degradation in reconstruction quality (25-30~dB PSNR drop) while preserving high imperceptibility (50-55~dB PSNR). Unlike pixel-space attacks, our method minimizes spectral artifacts, making detection more challenging. Across the Kodak, CLIC, and DIV2K datasets and three NIC models, attack impact is inversely related to image entropy: smoother images are more vulnerable. Weaker correlations on more diverse data (e.g., DIV2K) indicate additional complexity factors beyond texture. These results point to a broader vulnerability in VAE-based compression systems, which rely on entropy models that can be exploited through latent space perturbations.

From a practical standpoint, our findings suggest that NIC deployment in safety-critical or high-value pipelines (e.g., medical imaging, autonomous driving, commercial CDNs) requires new defense layers. Specifically, we recommend integrating wavelet-domain anomaly detection to flag high-frequency spectral irregularities and adopting entropy-adaptive coding schemes that allocate redundant bits to low-texture regions, thereby hardening the entropy bottleneck against targeted perturbations.

\emph{Future work} includes improving model robustness, studying transfer and query-based black-box variants of T-MLA across heterogeneous NIC architectures, and extending our approach to other Variational Autoencoder (VAE)-based systems such as video codecs, Neural Radiance Fields (NeRFs), and diffusion models. Our findings raise concerns about the security of generative compression pipelines and their broader deployment in modern AI systems.

\section*{Funding}
This work was supported by the Russian Science Foundation [grant number 25-41-00091]; and the National Natural Science Foundation of China (NSFC) [grant number W2412085]. The funders had no role in the study design; collection, analysis and interpretation of data; writing of the report; or the decision to submit the article for publication.

\section*{Data availability}
All datasets used in this study are publicly available (Kodak, CLIC-pro, DIV2K). 
Source code and experiment scripts to reproduce the results are available at \url{https://github.com/nkalmykovsk/tmla} (instructions are provided in the repository). No proprietary data were used.

\section*{Limitations}
We evaluate T-MLA in the white-box setting on three NICs; black-box/transfer variants and broader codec coverage (e.g., mbt2018) are deferred to future work. \autoref{sec:defense} outlines an initial wavelet-aware defense. Our goal is to provide a principled multiscale wavelet-domain framework that couples stealth and degradation, enabling reproducible robustness studies.

\section*{Declaration of generative AI and AI-assisted technologies in the manuscript preparation process}
During the preparation of this work the authors used ChatGPT (OpenAI) in order to polish grammar and phrasing, suggest literature-search keywords, and help debug minor Python issues in scripts. After using this service, the authors reviewed and edited the content as needed and take full responsibility for the content of the published article.

\bibliographystyle{model1-num-names}
\bibliography{references}

\clearpage
\appendix
\setcounter{table}{0}
\renewcommand{\thetable}{\Alph{section}.\arabic{table}}
\setcounter{figure}{0}
\renewcommand{\thefigure}{\Alph{section}.\arabic{figure}}

\section{Notations and Definitions}\label{sec:appendix}
\autoref{tab:notations} presents the complete list of variables, functions, and parameters with their corresponding definitions.

\begin{table}[h!]
  \centering
  \caption{Notations and definitions used in the paper.}
  \label{tab:notations}
  \begin{adjustbox}{max width=\columnwidth}
  \begin{tabular}{@{}l >{\raggedright\arraybackslash}p{0.7\columnwidth}@{}}
    \toprule
    \textbf{Notation} & \textbf{Definition} \\
    \midrule
    $\mathbf{x}$ & Original input image. \\
    $I,J,C$ & Image height, width, and number of channels. \\
    $f(\cdot;\theta)$ & Neural image compression (NIC) model with parameters $\theta$. \\
    $\hat{\mathbf{x}}$ & Reconstructed image, $\hat{\mathbf{x}} = f(\mathbf{x};\theta)$. \\
    $\mathbf{x}_{\mathrm{adv}}$ & Adversarial image (input image with perturbation). \\
    $\hat{\mathbf{x}}_{\mathrm{adv}}$ & Reconstructed adversarial image. \\
    $\mathcal{L}$ & Total optimization loss used to learn perturbations. \\
    $\mathcal{W}(\mathbf{x})$ & Wavelet decomposition, $\{ \mathbf{L}_S \} \cup \{ \mathbf{H}_k \}_{k=1}^S$. \\
    $\mathbf{W}_k$ & Subband at scale $k$; either $\mathbf{L}_S$ or one of $\mathbf{H}_k$. \\
    $\mathbf{L}_S$ & Low-frequency approximation subband at the coarsest scale. \\
    $\mathbf{H}_k$ & High-frequency detail subbands $\{\mathbf{LH}_k, \mathbf{HL}_k, \mathbf{HH}_k\}$. \\
    $S$ & Wavelet depth (number of DWT scales). \\
    $I_k,J_k$ & Spatial size of subbands at scale $k$. \\
    $\operatorname{DWT}^{(S)}$, $\operatorname{iDWT}^{(S)}$ & $S$-level discrete wavelet transform and its inverse. \\
    $\mathbf{N}_{\mathbf{W}}$ & Learnable noise tensor for a subband $\mathbf{W}$. \\
    $\mathbf{N}_k$ & Learnable noise tensor applied to subband at scale $k$. \\
    $\mathbf{n}$ & Spatial-domain additive noise, $\mathbf{x}_{\mathrm{adv}}=\mathbf{x}+\mathbf{n}$. \\
    $p(\mathbf{W},\mathbf{N})$ & Log--exp perturbation: $\operatorname{sign}(\mathbf{W})\log(\exp(|\mathbf{W}|)+\mathbf{N})$. \\
    $\delta$ & Global perturbation budget. \\
    $\delta_k$ & Per-subband budget, $\delta_k=\alpha^{S-k}\delta$. \\
    $\alpha$ & Scale adaptation factor for noise budget. \\
    $\eta$ & Learning rate for optimization. \\
    $\lambda$ & $\ell_1$ regularization weight for sparsity of $\{\mathbf{N}_k\}$. \\
    $Q_{\text{in}}, Q_{\text{out}}$ & Target PSNR for stealth (input) and degradation (output). \\
    $\mathbf{Q}$ & Tuple of targets $(Q_{\text{in}}, Q_{\text{out}})$. \\
    $\|\cdot\|_p,\,\|\cdot\|_\infty$ & Vector norms used for perturbation constraints. \\
    $\operatorname{clip}(\cdot)$ & Element-wise clipping to $[0,1]$. \\
    $\mathrm{PSNR},\,\mathrm{SSIM},\,\mathrm{VIF}$ & Quality metrics. \\
    $\mathrm{BPP}$ & Bits per pixel (bitrate). \\
    $E(x,y),\,\tilde{E}(x,y)$ & Local entropy and its normalized form used for complexity. \\
    $\mu_E$ & Mean normalized entropy over the image. \\
    $r$ & Radius of local neighborhood for entropy computation. \\
    \bottomrule
  \end{tabular}
  \end{adjustbox}
\end{table}
\FloatBarrier

\section{Image complexity characterization}
\label{sec:mean_entropy}

\begin{figure}
    \centering
    \begin{subfigure}[t]{0.48\linewidth}
        \centering
        \includegraphics[width=\linewidth]{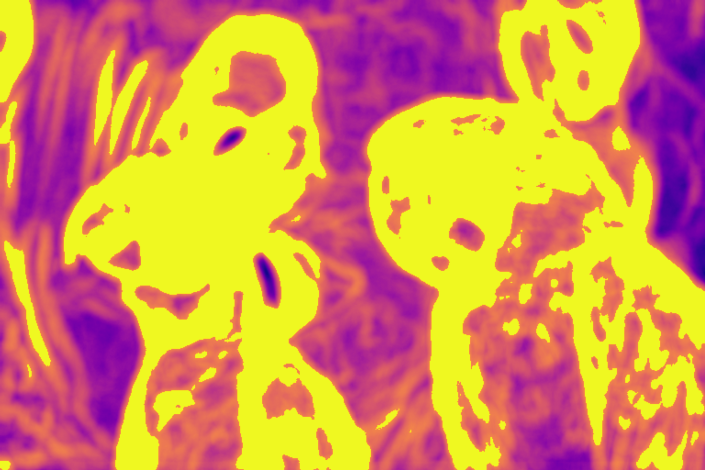}
        \caption{Entropy map showing texture complexity distribution.}
        \label{fig:entropy-map-a}
    \end{subfigure}
    \hfill
    \begin{subfigure}[t]{0.48\linewidth}
        \centering
        \includegraphics[width=\linewidth]{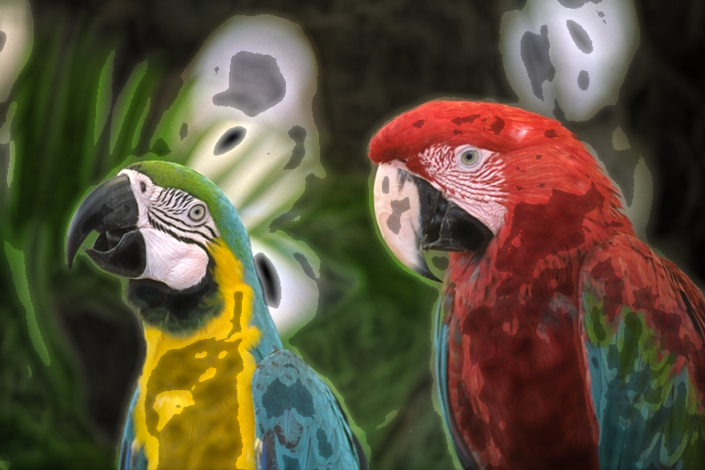}
        \caption{Entropy values overlaid on the original image.}
        \label{fig:entropy-map-b}
    \end{subfigure}
    \caption{
    Image complexity visualization using normalized local entropy. High-entropy regions (e.g., texture and detail) appear yellow; low-entropy (smooth, uniform) areas appear purple.
    }
    \label{fig:entropy-map}
\end{figure}

\subsection{Entropy Calculation}
\begin{figure}
    \centering
    \includegraphics[width=\columnwidth]{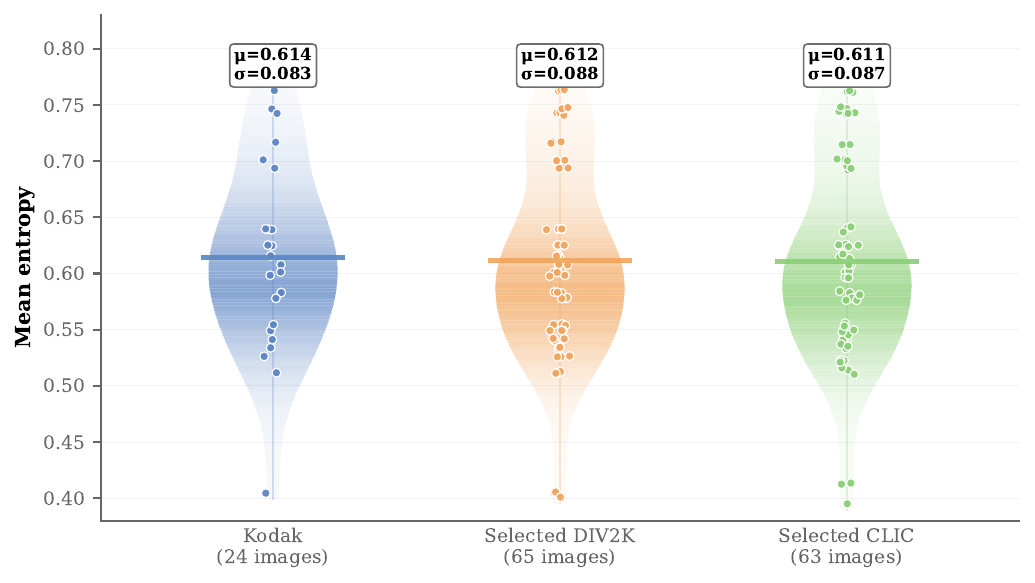}
    \caption{Distribution of mean entropy scores across test images from Kodak, DIV2K and CLIC datasets (from left to right). Higher entropy indicates greater textural complexity.}
    \label{fig:entropy-dist}
  \end{figure}

We quantify image complexity using normalized local entropy. Given an RGB image $\mathbf{x} \in [0,1]^{H \times W \times 3}$, we convert it to grayscale $Y \in \{0,\ldots,255\}^{H \times W}$ and compute local Shannon entropy in a circular neighborhood $\mathcal{N}_r(x,y)$ of radius $r=10$ (empirically balances locality and statistical stability):
\begin{align}
  E(x,y) &= -\sum_{i=0}^{255} p_i(x,y) \log_2 p_i(x,y),\\
  p_i(x,y) &= \frac{|\{(u,v) \in \mathcal{N}_r(x,y) : Y_{u,v}=i\}|}{|\mathcal{N}_r(x,y)|}.
\end{align}
The entropy values are normalized by $E_{\max} = 8$ bits to obtain $\tilde{E}(x,y) = E(x,y)/E_{\max} \in [0,1]$ (see \autoref{fig:entropy-map}). Finally, the global complexity score is obtained by averaging normalized entropy values across all spatial locations:
\[
\mu_E = \frac{1}{H \cdot W}\sum_{x=1}^{H}\sum_{y=1}^{W}\tilde{E}(x,y).
\]
For our experiments, this metric ranges from 0.4 (uniform regions) to 0.78 (highly textured areas) on our evaluation images (see \autoref{fig:entropy-dist}).

\subsection{Full Correlation Analysis}

\begin{figure*}[t]
    \centering
    \includegraphics[width=0.96\textwidth]{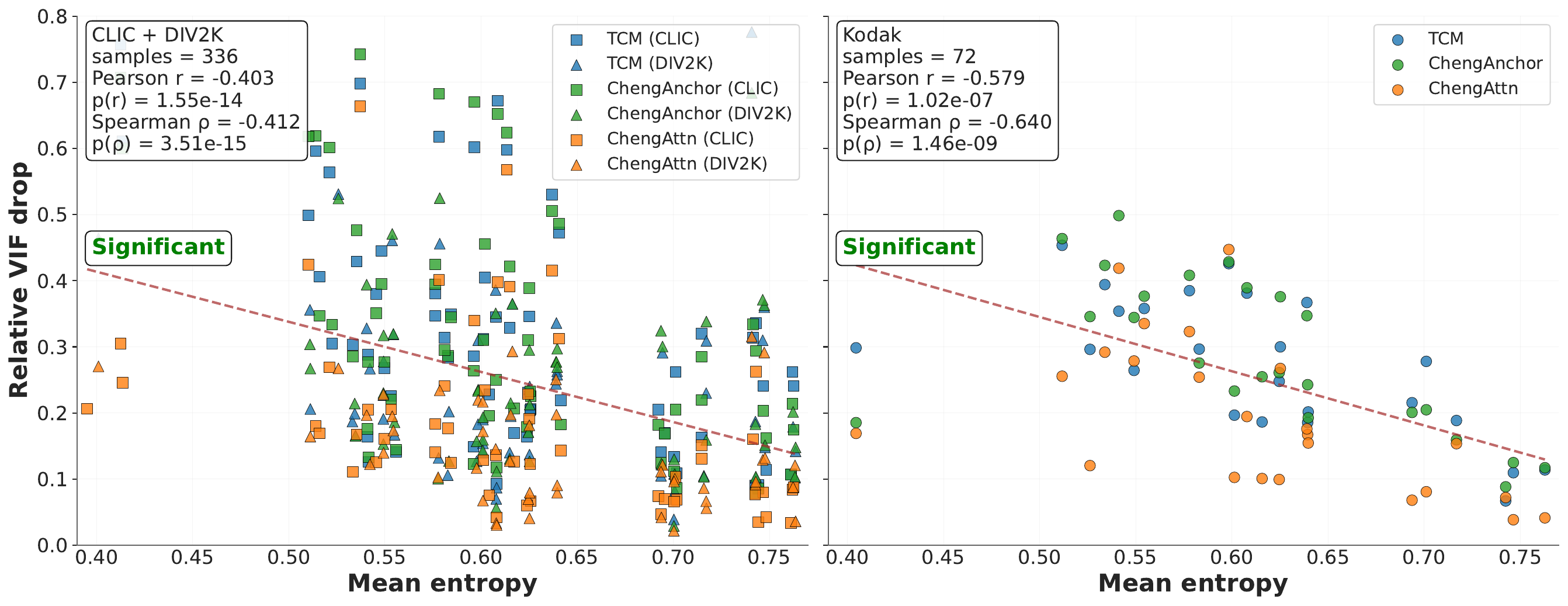}
    \caption{
    The full correlation between mean entropy and relative VIF drop across Kodak, DIV2K and CLIC images for three models Cheng2020-Anchor, Cheng2020-Attention and LIC-TCM. Lower entropy images suffer greater perceptual quality loss under attack, revealing a content-aware vulnerability.
    }
    \label{fig:full_correlation}
\end{figure*}
\begin{figure*}[h!]
    \centering
    \includegraphics[width=\linewidth]{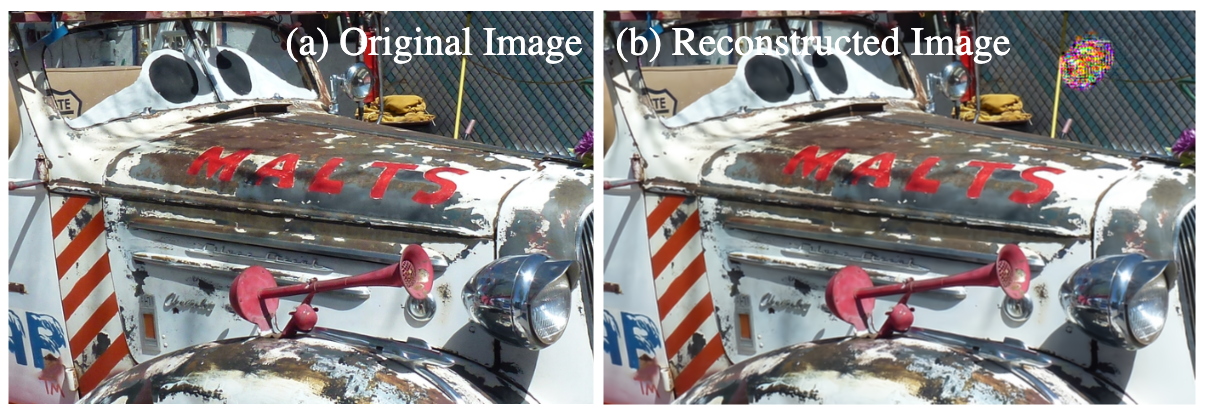}
    \caption{Example of DIV657 dataset.}
    \label{fig:example-div-657}
\end{figure*}
\begin{figure*}[h!]
    \centering
    \includegraphics[width=\linewidth]{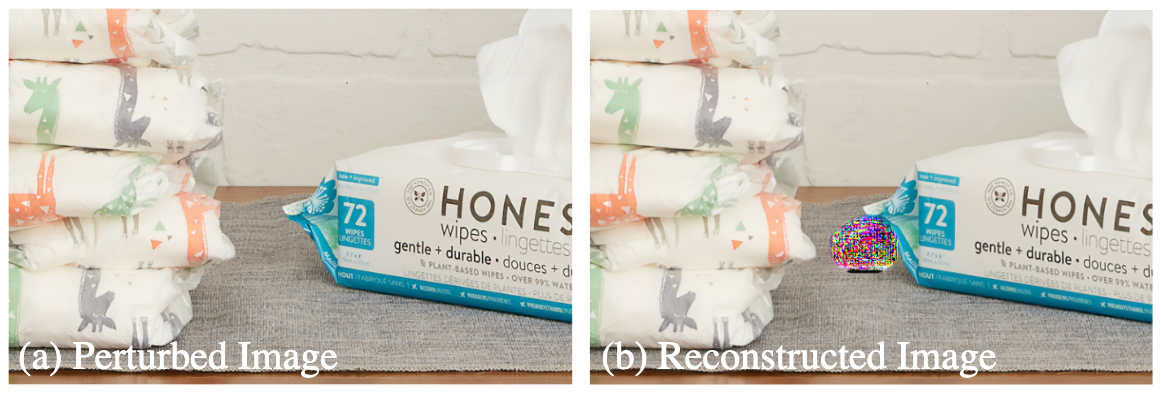}
    \caption{Example of DIV155 dataset.}
    \label{fig:example-div-155}
\end{figure*}
\begin{figure*}[h!]
    \centering
    \includegraphics[width=\linewidth]{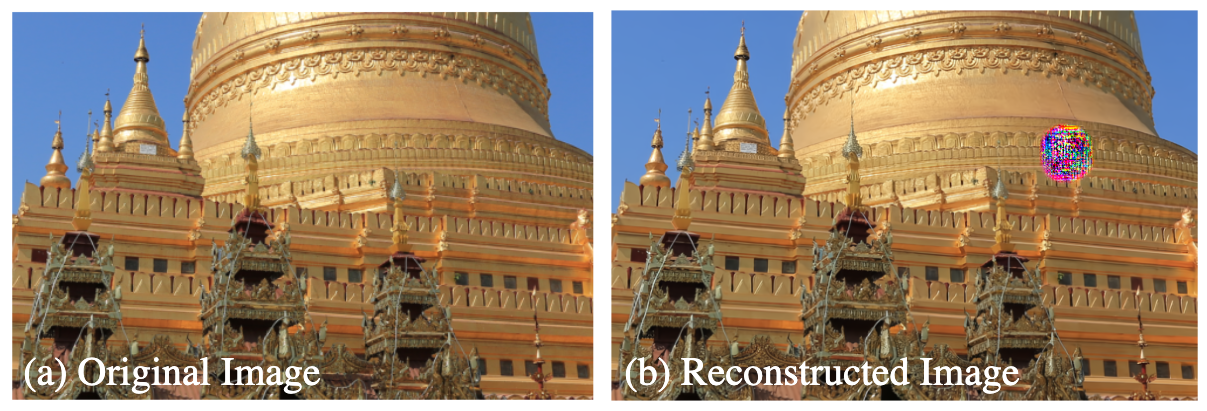}
    \caption{Example of DIV049 dataset.}
    \label{fig:example-div-049}
\end{figure*}

\autoref{fig:full_correlation} shows correlation analysis between mean entropy and relative VIF drop across all datasets and compression models. For DIV2K and CLIC datasets, we excluded $10\%$ of samples with mean entropy $< 0.6$ and relative VIF drop $< 0.1$. This filtering removes outliers with significant variations in depth maps and non-standard texture distributions (see examples in \autoref{fig:example-div-657}, \autoref{fig:example-div-155}, and \autoref{fig:example-div-049}). Crucially, this selection ensures that the complexity distribution of the larger datasets aligns with the standard Kodak benchmark, preventing results from being skewed by trivial, low-information samples and ensuring generalization to realistic, detail-rich content.

The consistent negative correlation across all models confirms the entropy vulnerability relationship. Cheng2020-Attention (marked in orange in \autoref{fig:full_correlation}) exhibits weaker correlations than other architectures, suggesting enhanced robustness.

Dataset variations reveal factors beyond entropy affecting model vulnerability: Kodak shows the strongest correlations ($r = -0.579$), while DIV2K and CLIC show weaker correlations ($r = -0.241$). These differences stem from dataset characteristics: Kodak contains professionally captured photographs with controlled lighting and predictable texture distributions, while DIV2K and CLIC feature more diverse natural images and compression-optimized content, respectively.

\section{Multiscale Haar Wavelet Decomposition}
\label{sec:wavelet}

\subsection{Intuition}
The \emph{Haar} transform is the simplest orthonormal wavelet.  
At each scale, it separates an image into one \emph{approximation band} (low–frequency content) and three \emph{detail bands} that capture horizontal, vertical, and diagonal edges.  
This is achieved by filtering every non-overlapping \(2\times2\) pixel block with the 1-D low-/high-pass filters  
\(h=[\tfrac{1}{\sqrt{2}},\tfrac{1}{\sqrt{2}}]\) and \(g=[\tfrac{1}{\sqrt{2}},-\tfrac{1}{\sqrt{2}}]\) along both spatial dimensions.  
For clarity, we write the resulting 2-D formulas with the equivalent factor \(\tfrac{1}{2}=1/(\sqrt{2}\!\cdot\!\sqrt{2})\).

\subsection{Forward transform}
Let
\begin{equation}
\begin{bmatrix}
I_{2i,2j} & I_{2i,2j+1}\\
I_{2i+1,2j} & I_{2i+1,2j+1}
\end{bmatrix}
\end{equation}
be a block centered at coordinates \((2i,2j)\).
The four subband coefficients are as follows:
\begin{align*}
LL_{i,j} &= \tfrac{1}{2}\!\bigl(I_{2i,2j}+I_{2i,2j+1}+I_{2i+1,2j}+I_{2i+1,2j+1}\bigr),\\[2pt]
LH_{i,j} &= \tfrac{1}{2}\!\bigl(I_{2i,2j}-I_{2i,2j+1}+I_{2i+1,2j}-I_{2i+1,2j+1}\bigr),\\[2pt]
HL_{i,j} &= \tfrac{1}{2}\!\bigl(I_{2i,2j}+I_{2i,2j+1}-I_{2i+1,2j}-I_{2i+1,2j+1}\bigr),\\[2pt]
HH_{i,j} &= \tfrac{1}{2}\!\bigl(I_{2i,2j}-I_{2i,2j+1}-I_{2i+1,2j}+I_{2i+1,2j+1}\bigr)
\end{align*}

\subsection{Inverse transform (reconstruction)}
Because the transform is orthonormal, the inverse is obtained by simple addition of the detail bands to the approximation band:
\begin{align*}
I_{2i,2j}      &= LL_{i,j}+LH_{i,j}+HL_{i,j}+HH_{i,j},\\
I_{2i,2j+1}    &= LL_{i,j}-LH_{i,j}+HL_{i,j}-HH_{i,j},\\
I_{2i+1,2j}    &= LL_{i,j}+LH_{i,j}-HL_{i,j}-HH_{i,j},\\
I_{2i+1,2j+1}  &= LL_{i,j}-LH_{i,j}-HL_{i,j}+HH_{i,j}
\end{align*}

\subsection{Recursive decomposition}
Applying the same procedure to the approximation band \(LL\) yields a multiresolution pyramid.  
For a \(768\times512\) input, we use three levels (as in our experiments):
\begin{center}
\setlength{\tabcolsep}{4pt}
\begin{tabular}{@{}c@{\hspace{6pt}}c@{\hspace{6pt}}c@{}}
\textbf{Level} & \textbf{Bands} & \textbf{Size}\\
\hline
1 & $LL^{(1)},LH^{(1)},HL^{(1)},HH^{(1)}$ & $384{\times}256$\\
2 & $LL^{(2)},LH^{(2)},HL^{(2)},HH^{(2)}$ & $192{\times}128$\\
3 & $LL^{(3)},LH^{(3)},HL^{(3)},HH^{(3)}$ & $96{\times}64$
\end{tabular}
\end{center}

\subsection{Relation to our attack}
Natural images exhibit rapidly decaying energy in the detail bands; most coefficients are near zero. Our log–exponential perturbation strategy injects noise into \emph{ten} subbands simultaneously (all detail bands and only the final approximation band $LL^{(3)}$ due to the pyramidal dependency), deliberately deviating from this sparsity pattern. This targeted high-frequency perturbation approach makes the noise particularly difficult for the human eyes to detect, while maximizing downstream distortion.

\subsection{Scale ablation}
\begin{figure*}[t]
    \centering
  
    \begin{subfigure}{\textwidth}
      \centering
      \includegraphics[width=\textwidth]{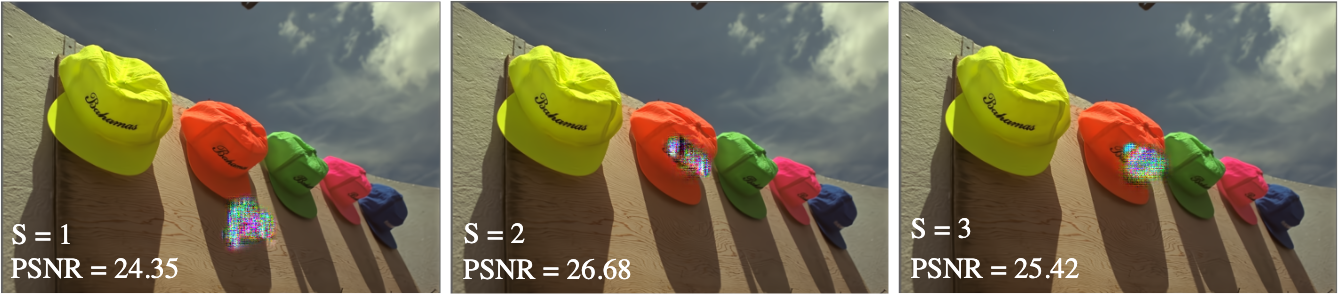}
      \subcaption{Visual comparison of attack results with different wavelet decomposition scales. The PSNR values between perturbed images and attacked outputs after NIC are similar across scales. However, for $S=1$ the perturbations are concentrated in flat regions (wall), while for higher scales they appear in more textured areas (text on the cap).}
      \label{fig:scales-a}
    \end{subfigure}
  
    \medskip
  
    \begin{subfigure}{\textwidth}
      \centering
      \includegraphics[width=\textwidth]{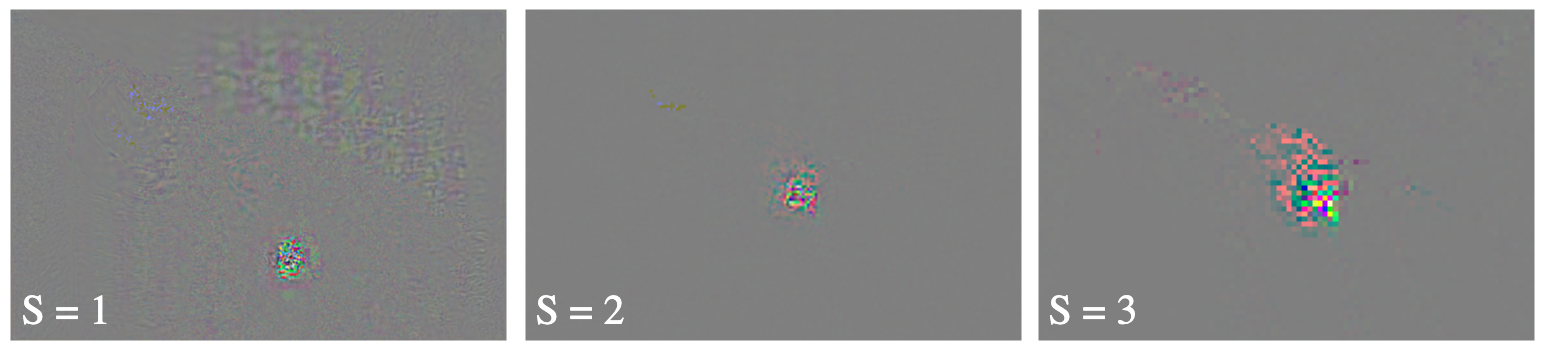}
      \subcaption{Difference maps between original and perturbed images at different scales: (a) scale=1, (b) scale=2, and (c) scale=3, illustrating how perturbation patterns vary across wavelet decomposition levels.}
      \label{fig:scales-b}
    \end{subfigure}
  
    \caption{Comparison across wavelet scales: (a) qualitative results; (b) difference maps.}
    \label{fig:scales-combined}
  \end{figure*}

We conducted ablation experiments with different scale levels ($S = 1, 2, 3$) to analyze the impact of multiscale perturbations in our attack framework. Our key findings were:

\begin{itemize}
    \item \textbf{Quantitative Results:} PSNR metrics (\autoref{tab:scale_ablation}) remained comparable across all scale configurations, suggesting similar overall perturbation magnitudes.

    \item \textbf{Single-scale Attack ($S=1$):} Noise distributed more uniformly, including flat image regions (\autoref{fig:scales-a}), making perturbations more visually apparent despite similar PSNR values. 

    \item \textbf{Multiscale Attacks ($S=2,3$):} Perturbations naturally concentrated in textured image regions (\autoref{fig:scales-a}), significantly reducing human perceptibility. This validates our choice of $S=3$ for the main experiments.
\end{itemize}

Finally, the difference maps between original and perturbed images for different scales (\autoref{fig:scales-b}) further illustrate how the perturbation patterns vary across wavelet decomposition levels.

\begin{table}[t]
\centering
\caption{Ablation study on different Haar wavelet decomposition scales}
\label{tab:scale_ablation}
\begin{tabular}{ccccccc}
\toprule
\multirow{2}{*}{Scale} & \multicolumn{2}{c}{Stealth $\uparrow$} & \multicolumn{4}{c}{Attack Success $\downarrow$} \\
\cmidrule(lr){2-3} \cmidrule(lr){4-7}
 & PSNR & VIF & PSNR & SSIM & VIF & BPP \\
\midrule
S=1 & 55.1 & 0.999 & 24.4 & 0.972 & 0.548 & 0.47 \\
S=2 & 54.3 & 0.998 & 26.7 & 0.975 & 0.683 & 0.47 \\
S=3 & 54.8 & 0.998 & 25.4 & 0.974 & 0.624 & 0.58 \\
\bottomrule
\end{tabular}
\end{table}

\section{Wavelet family}
\label{sec:wavelet_families}

We evaluated several wavelet families for adversarial attacks against NIC models (see \autoref{tab:wavelet_families} for detailed characteristics):

\begin{itemize}
\item \textbf{Haar}: The simplest orthogonal wavelet with compact support, efficient to compute.
\item \textbf{Biorthogonal 1.1} (Bior1.1): Symmetric wavelet filters with exact reconstruction properties.
\item \textbf{Coiflets 2} (Coif2): Both scaling and wavelet functions have vanishing moments, improving edge representation.
\item \textbf{Daubechies 2} (DB2): Orthogonal wavelets with excellent energy compaction for natural textures.
\item \textbf{Symlets 2} (Sym2): Near-symmetric variant of Daubechies wavelets, reducing shift sensitivity.
\end{itemize}

\begin{figure*}[h!]
    \centering
    \includegraphics[width=\textwidth]{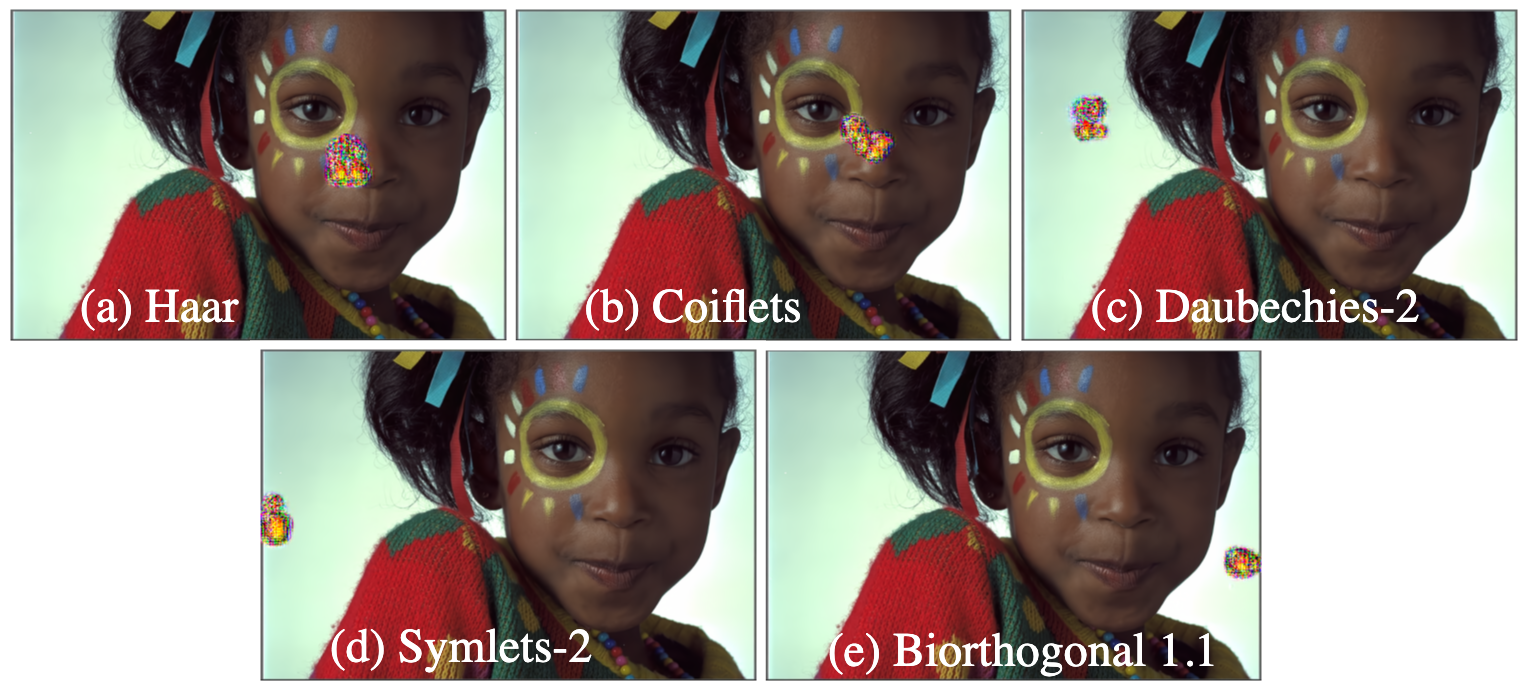}
    \caption{Visual comparison of different wavelet families for reconstructed images after targeted multiscale log-exponential adversarial attack (\textbf{T-MLA}).}
    \label{fig:wavelet_families}
\end{figure*}

The results show comparable performance across all wavelet families, with PSNR values ranging from 53.9 to 55.1 dB for stealth and 24.7 to 25.3 dB for attack success. Given this similarity in effectiveness, we recommend using Haar wavelets due to their computational efficiency and implementation simplicity. For visual comparison, see \autoref{fig:wavelet_families}.

\begin{table}[t]
    \centering
    \caption{Comparison of different wavelet families for adversarial attacks}
    \label{tab:wavelet_families}
    \begin{tabular}{ccccccc}
    \toprule
    \multirow{2}{*}{Family} & \multicolumn{2}{c}{Stealth $\uparrow$} & \multicolumn{4}{c}{Attack Success $\downarrow$} \\
    \cmidrule(lr){2-3} \cmidrule(lr){4-7}
     & PSNR & VIF & PSNR & SSIM & VIF & BPP \\
    \midrule
    Haar    & 55.1 & 0.999 & 24.7 & 0.962 & 0.653 & 0.65 \\
    Bior1.1 & 53.9 & 0.999 & 25.3 & 0.965 & 0.799 & 0.65 \\
    Coif2   & 55.0 & 0.999 & 25.3 & 0.962 & 0.690 & 0.65 \\
    DB2     & 55.0 & 0.999 & 25.0 & 0.963 & 0.747 & 0.65 \\
    Sym2    & 54.0 & 0.999 & 25.0 & 0.965 & 0.789 & 0.65 \\
    \bottomrule
    \end{tabular}
    \end{table}

\section{Empirical Estimation of Wavelet Energy Decay Across Scales}
\label{sec:empirical-alpha}

To justify the adaptive scaling factor \( \alpha \) used in our subband-specific noise bounds, we analyze the empirical decay in wavelet coefficient magnitudes across scales using natural images from the DIV2K and CLIC validation datasets.

Let \( \mathcal{W}(\mathbf{x}) = \{ \mathbf{L}_S, \mathbf{H}_S, \dots, \mathbf{H}_1 \} \) denote the $S$-level discrete wavelet transform (DWT) of an image \( \mathbf{x} \), where \( \mathbf{H}_S = \{ \mathbf{LH}_S, \mathbf{HL}_S, \mathbf{HH}_S \} \) are the high-frequency detail subbands at scale \( S \), and \( \mathbf{L}_S \) is the coarsest approximation band.

For each image, we compute the maximum absolute coefficient in each subband, which we denote \( m_k \), where \( k \) indexes the subbands ordered from finest to coarsest scale. We then define the per-image ratio between adjacent subbands:
\[
r_k = \frac{m_{k+1}}{m_k}.
\]

Given a dataset \( \mathcal{D} = \{ \mathbf{x}_1, \dots, \mathbf{x}_N \} \), we compute the median ratio across subbands for each image:
\[
\rho_i = \text{median}(r_1^{(i)}, \dots, r_S^{(i)}),
\]
and define the empirical average scaling factor as:
\[
\bar{\alpha}_{\text{emp}} = \frac{1}{N} \sum_{i=1}^N \rho_i.
\]

We apply this method to Haar-DWT decompositions of all images in the DIV2K and CLIC validation sets using a 5-level decomposition (\( S = 5 \)). For each subband, we take the per-channel maximum across absolute wavelet coefficients, then compute the per-scale ratios and aggregate as above.

\subsection{Result} The empirical scaling factor is:
\[
\bar{\alpha}_{\text{emp}} \approx 1.9
\]
This value closely matches the theoretical expectation of \( \alpha = 2 \) under the assumption of energy doubling across scales. In practice, we adopt \( \alpha = 1.8 \) to ensure a slightly more conservative bound, which improves perceptual quality without sacrificing attack strength.

\subsection{Implementation details} For each image \( \mathbf{x} \), we decompose it into subbands using multilevel Haar wavelets. We then extract:
\begin{itemize}
    \item The maximum absolute value from each subband.
    \item The scale-wise ratios between these maximums.
    \item The per-image median of the ratios.
\end{itemize}
Finally, we average over all images to obtain \( \bar{\alpha}_{\text{emp}} \).

\section{Defense Studies}
\label{sec:defense}

The defense algorithm is conceptually similar to the attack algorithm in the main paper, with a key difference: the attack maximizes distortion using gradient ascent, while the defense minimizes distortion using gradient descent.

\begin{figure*}[t]
    \centering
    \includegraphics[width=\linewidth]{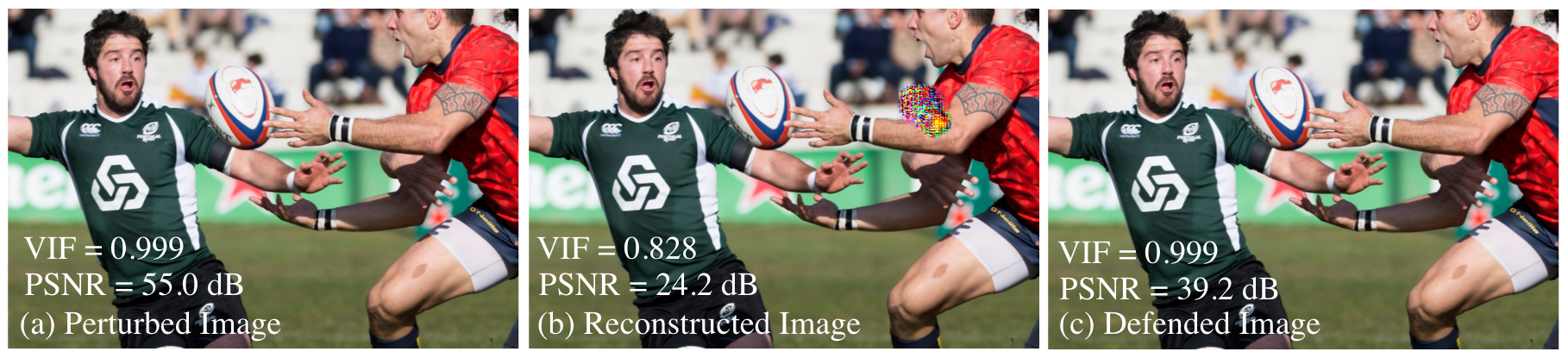}
    \caption{
    Defended image for DIV090 image. 
    (a) The adversarial perturbed image ($\mathrm{PSNR}=55.0$~dB, $\mathrm{VIF}=0.999$) looks clean to the human eye. 
    (b) The attacked reconstruction ($\mathrm{PSNR}=24.2$~dB, $\mathrm{VIF}=0.828$) exhibits visible artifacts, particularly on the player's arm. 
    (c) The defended image ($\mathrm{PSNR}=39.2$~dB, $\mathrm{VIF}=0.999$) obtained via our iterative defense (\autoref{alg:defense}) restores quality: while PSNR is lower than the stealthy input, it is higher than the attacked output, and artifacts are visually removed.
    }
    \label{fig:defended-div-090}
\end{figure*}

\begin{algorithm}[t]
    \caption{Adaptive Defense Against \textbf{T-MLA}}
    \label{alg:defense}
    \begin{algorithmic}[1]
    \Require{Input image $\mathbf{x}$, NIC model $f(\cdot;\theta)$, initial noise budget $\delta$,
             perceptual loss $\mathcal{L}$, learning rate $\eta$,
             iterations $T$}
    \Ensure{Defended image $\mathbf{x}^{\star}$, reconstructed output $\hat{\mathbf{x}}^{\star}$, defense noise $\mathbf{n}^{\star}$}
    \State Initialize $\mathbf{n}^{(0)}\!\sim\!\mathcal{N}(0,\,\delta^{2})$
    \For{$t = 0$ to $T-1$}
        \State $\mathbf{n}^{\text{clip}}\!\gets\!\text{clip}(\mathbf{n},\,-\delta,\,\delta)$ \Comment{amplitude constraint}
        \State $\mathbf{x}'\!\gets\!\mathbf{x} + \mathbf{n}^{\text{clip}}$ \Comment{Apply defense perturbation}
        \State $\hat{\mathbf{x}} \gets f(\mathbf{x}';\theta)$ \Comment{Forward pass through codec}
        \State $\ell \gets \mathcal{L}(\hat{\mathbf{x}},\,\mathbf{x})$ \Comment{Total objective}
        \State $\mathbf{n}^{(t\!+\!1)} \gets \mathbf{n}^{(t)} + \eta\,\nabla_{\mathbf{n}^{(t)}}\ell$ \Comment{Gradient update}
    \EndFor
    \State $\mathbf{x}^{\star} \gets \mathbf{x} + \mathbf{n}^{\star}$ \Comment{Apply defense}
    \State $\hat{\mathbf{x}}^{\star} \gets f(\mathbf{x}^{\star};\theta)$ \Comment{Final reconstruction}
    \State \Return $\mathbf{x}^{\star}$, $\hat{\mathbf{x}}^{\star}$, $\mathbf{n}^{\star}$
    \end{algorithmic}
\end{algorithm}

\subsection{Algorithm Overview}
The proposed defense adaptively learns a frequency-aware perturbation to neutralize adversarial attacks while preserving legitimate image content. Key innovations include:

\begin{itemize}
    \item \textbf{Multiscale frequency analysis:} Uses wavelet decomposition to enforce sparsity constraints across different frequency bands, preventing over-smoothing of fine details.
    
    \item \textbf{Adaptive noise scheduling:} Gradually reduces the perturbation budget to find minimal defensive modifications that restore compression fidelity.
    
    \item \textbf{Perceptual optimization:} Supports multiple fidelity metrics to balance reconstruction quality against visual artifacts.
\end{itemize}

An example of our defense approach can be seen in \autoref{fig:defended-div-090}, which shows: (a) the perturbed image containing adversarial noise, (b) the reconstructed output after compression showing significant distortion, and (c) the defended image where our method successfully mitigates the attack while preserving visual quality. The defense effectively removes adversarial artifacts while maintaining the natural image content and texture details.

\section{Convergence and Sensitivity Analysis}
\label{sec:convergence_analysis}

\begin{figure*}[t]
    \centering
    \begin{subfigure}[b]{0.48\linewidth}
        \centering
        \includegraphics[width=\linewidth]{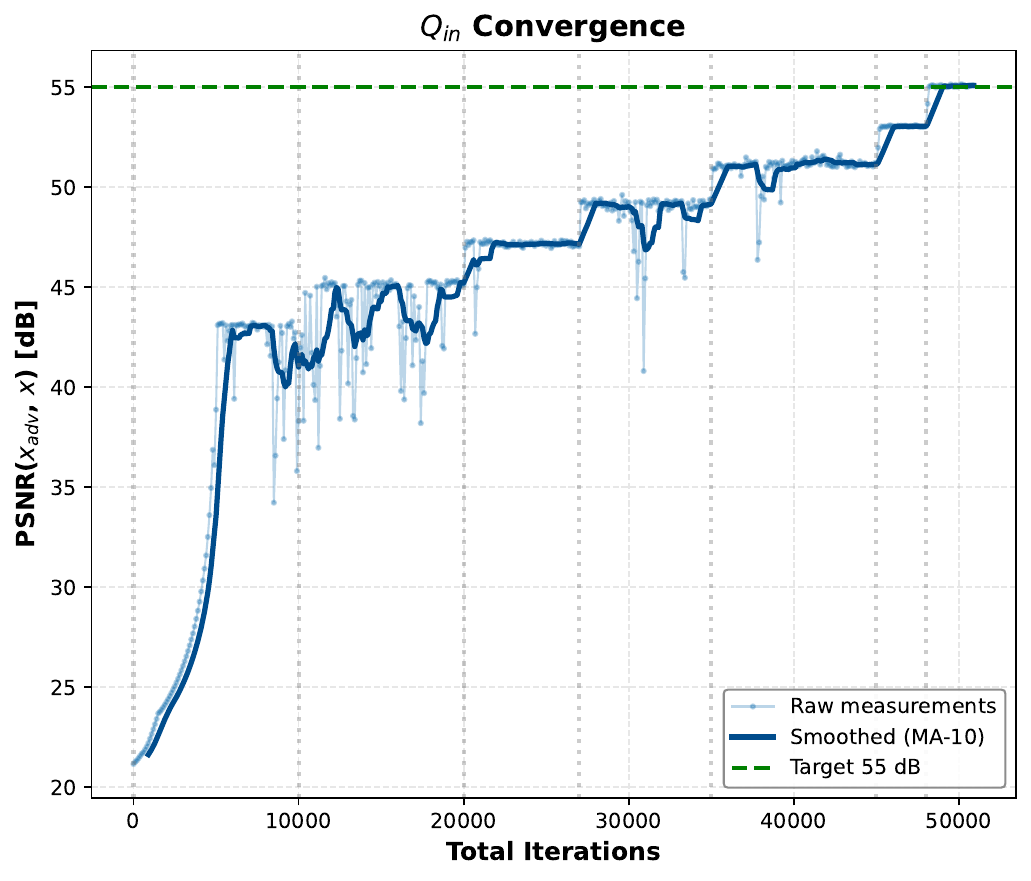}
        \caption{Input PSNR $Q_{\text{in}}$.}
        \label{fig:qin_convergence}
    \end{subfigure}
    \hfill
    \begin{subfigure}[b]{0.48\linewidth}
        \centering
        \includegraphics[width=\linewidth]{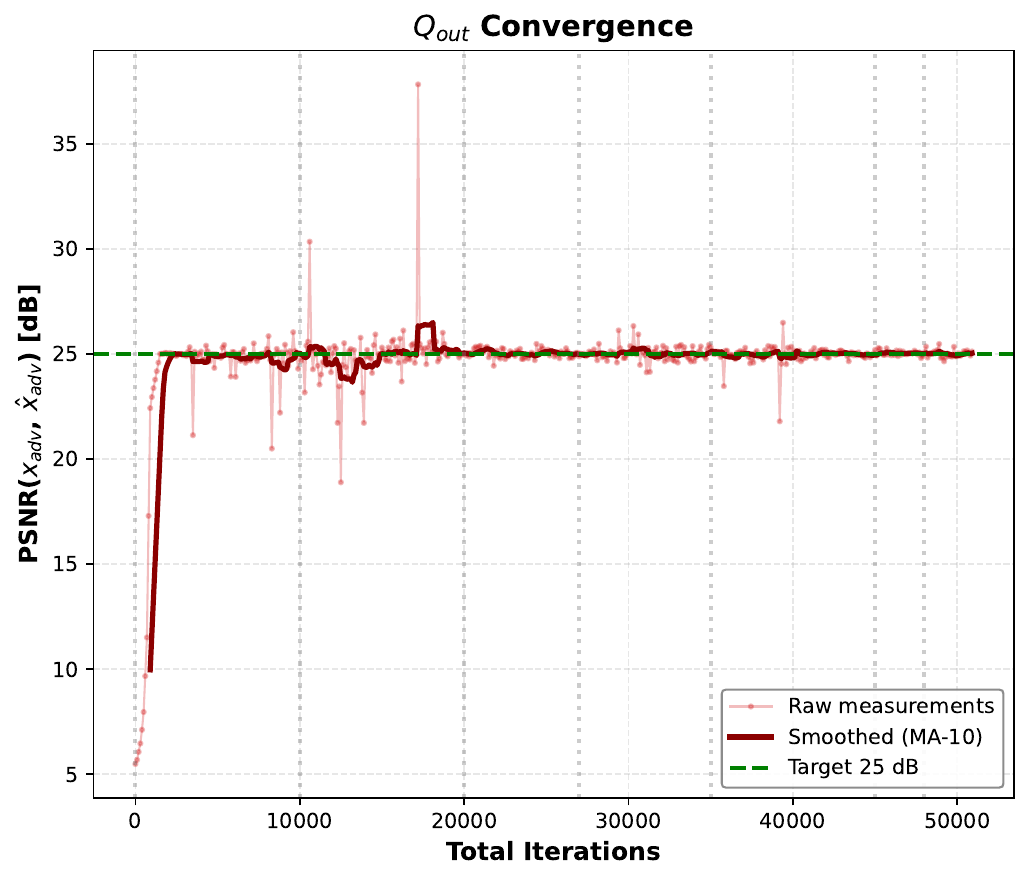}
        \caption{Output PSNR $Q_{\text{out}}$.}
        \label{fig:qout_convergence}
    \end{subfigure}
    \caption{
    Convergence of T-MLA on Kodak23 with Cheng2020-Anchor ($S{=}3$ scales, $Q_{\text{in}}{=}55$~dB, $Q_{\text{out}}{=}25$~dB).
    The input PSNR (left) progressively increases towards the target while the output PSNR (right) remains constrained near 25~dB, illustrating stable optimization under the joint stealth--degradation objective.
    }
    \label{fig:convergence}
\end{figure*}

We analyze the convergence behavior of T-MLA on a representative case (kodak23, Cheng2020-Anchor, $S{=}3$, $Q_{\text{in}}{=}55$~dB, $Q_{\text{out}}{=}25$~dB). As shown in \autoref{fig:convergence}, the input PSNR $Q_{\text{in}}$ increases in stages: the optimizer reaches an intermediate PSNR level, briefly stabilizes, and then moves to the next, tighter constraint until converging near 55~dB. Throughout optimization, the output PSNR $Q_{\text{out}}$ remains concentrated around the 25~dB target, indicating that the joint stealth--degradation objective is enforced reliably.

Regarding hyperparameters, we empirically select the largest feasible $Q_{\text{in}}$ and the smallest feasible $Q_{\text{out}}$ for which optimization converges. Very low $Q_{\text{out}}$ targets (extreme degradation) or excessively large scale factors $\alpha$ shrink the feasible perturbation region and can destabilize convergence; in practice, small changes of $\alpha$ (e.g., $\pm 0.1$ around $1.8$) do not noticeably affect convergence. In the main experiments we fix $(Q_{\text{in}},Q_{\text{out}})=(50\text{--}55,25)$~dB, $\alpha = 1.8$, and a moderate sparsity weight $\lambda$; in contrast to $\alpha$, the method is sensitive to $\lambda$, and even changes at the third decimal place can alter convergence behavior, with too large $\lambda$ enforcing overly sparse perturbations and degrading convergence.

\section{Perceptual Robustness Evaluation}
\label{sec:perceptual_robustness}

\begin{figure*}[t]
    \centering
    \includegraphics[width=0.96\textwidth]{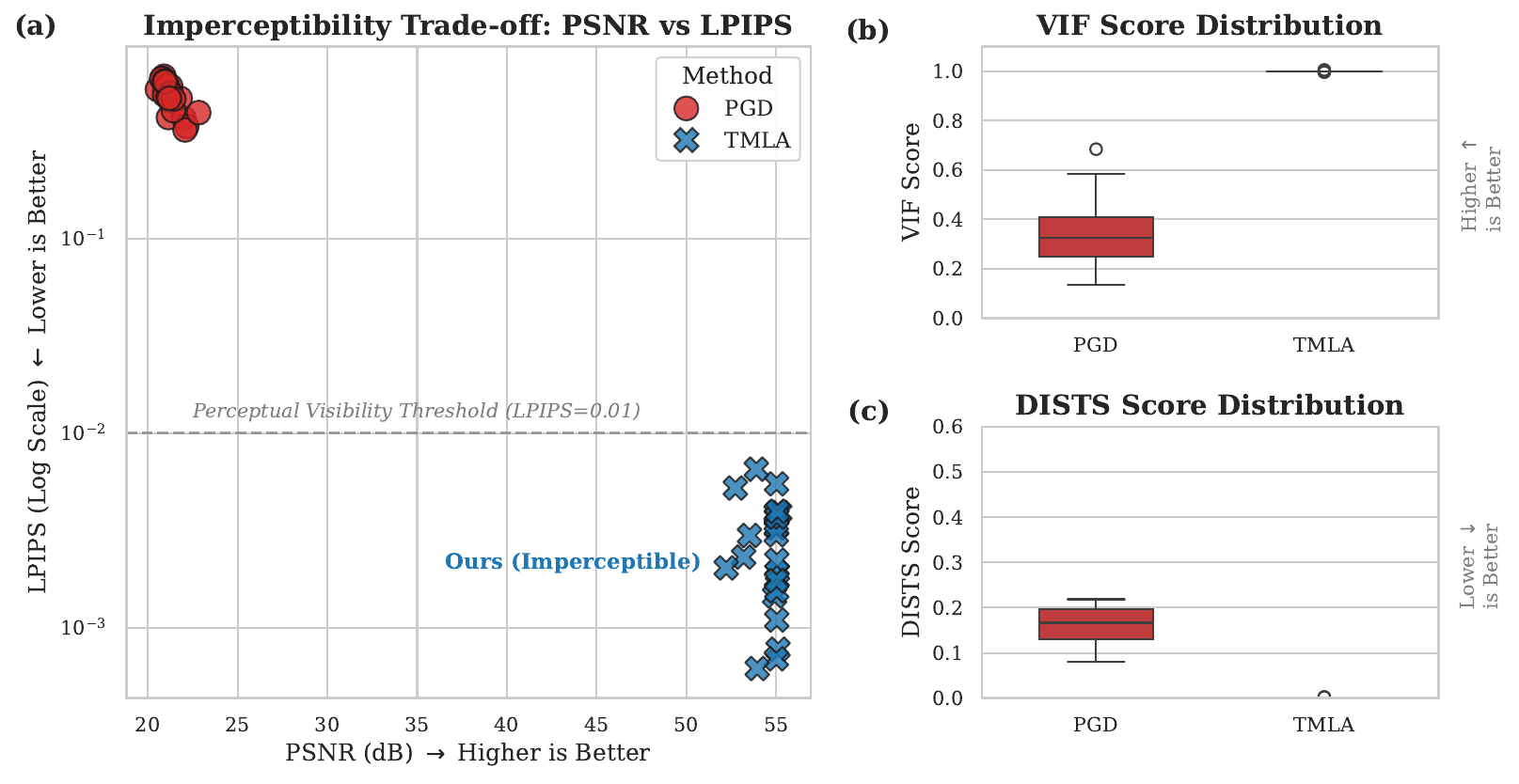}
    \caption{
    Perceptual robustness evaluation on Cheng2020-Anchor across Kodak dataset.
    (a) Trade-off between LPIPS (log scale) and PSNR: T-MLA (blue crosses) stays well below the visibility threshold ($0.01$) with high PSNR, unlike PGD (red circles).
    (b) VIF score distribution: T-MLA maintains values near 1.0 (best).
    (c) DISTS score distribution: T-MLA yields scores near 0 (best), indicating minimal texture distortion compared to PGD.
    }
    \label{fig:imperceptibility}
\end{figure*}

To validate the perceptual invisibility of our adversarial perturbations, we extended our evaluation beyond standard MSE-based metrics (PSNR, SSIM) to include deep learning-based perceptual metrics: LPIPS (Learned Perceptual Image Patch Similarity) and DISTS (Deep Image Structure and Texture Similarity).

As illustrated in \autoref{fig:imperceptibility}, our method achieves a mean LPIPS score of 0.0028$\pm$0.0016 on the Kodak dataset with Cheng2020-Anchor. This is nearly two orders of magnitude lower than the PGD baseline (0.545) and consistently remains well below the widely accepted visibility threshold of 0.01 (maximum observed score: 0.0065). Similarly, for DISTS, T-MLA attains a near-perfect mean score of 0.0008, compared to 0.161 for the baseline.

While we observe that PSNR and LPIPS do not perfectly correlate in the high-quality regime ($PSNR > 50$ dB) due to the sensitivity of deep metrics to microscopic noise, the global trend is consistent: the massive PSNR gain (+33.3 dB over baseline) translates directly into imperceptible LPIPS/DISTS scores, confirming that our perturbations are perceptually invisible.

\section{Additional Results}
\label{sec:results}

In this section, we provide detailed quantitative results in \autoref{tab:chenganchor_results_kodak}, \autoref{tab:chenganchor_results_clic}, \autoref{tab:chenganchor_results_div2k}, \autoref{tab:chenganchor_results_div2k_stats}, \autoref{tab:chengattn_results_kodak}, \autoref{tab:chengattn_results_clic}, \autoref{tab:chengattn_results_div2k}, \autoref{tab:chengattn_results_div2k_stats}, \autoref{tab:tcm_results_kodak}, \autoref{tab:tcm_results_clic}, \autoref{tab:tcm_results_div2k}.  These tables present comprehensive per-image attack performance metrics for three models (Cheng2020-Anchor, Cheng2020-Attention, and LIC-TCM) evaluated on Kodak, CLIC, and DIV2K datasets. For each image, we report:
stealth metrics ($\uparrow$): PSNR and VIF scores measuring imperceptibility; attack success metrics ($\downarrow$): PSNR, SSIM, and VIF measuring distortion; compression rate in bits-per-pixel (BPP).

\begin{figure*}[t]
    \centering
    \includegraphics[width=\textwidth]{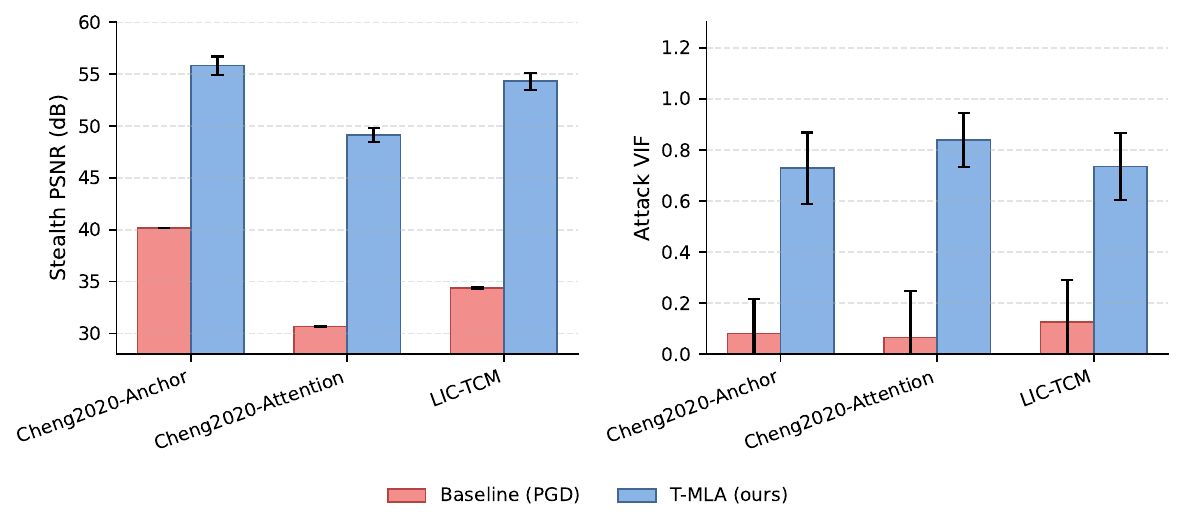}
    \caption{
    Statistical comparison between the baseline PGD attack and our T-MLA on the Kodak dataset (24 images).
    Bars show mean $\pm$ standard deviation of Stealth PSNR (left, higher is better)
    and Attack VIF (right, lower is better) for Cheng2020-Anchor, Cheng2020-Attention, and LIC-TCM.
    All improvements of T-MLA over PGD are statistically significant according to a two-sided paired t-test on per-image metrics ($p < 10^{-8}$ for all cases).
    }
    \label{fig:tmla_vs_pgd_kodak}
\end{figure*}

\begin{table*}[t]
  \centering
  \caption{Summary statistics of different attack methods for \textit{Cheng2020-Anchor} on the Kodak dataset (24 images).}
  \label{tab:apgd_anchor_kodak_stats}
  {\small\setlength{\tabcolsep}{12pt}
  \begin{adjustbox}{max width=\textwidth}
  \begin{tabular}{@{}lcccccc@{}}
    \toprule
    \textbf{Method}
    & \multicolumn{2}{c}{\textbf{Stealth} $\uparrow$}
    & \multicolumn{4}{c}{\textbf{Attack Success} $\downarrow$} \\
    \cmidrule(lr){2-3} \cmidrule(lr){4-7}
    & \textbf{PSNR}
    & \textbf{VIF}
    & \textbf{PSNR}
    & \textbf{SSIM}
    & \textbf{VIF}
    & \textbf{BPP} \\
    \midrule
    PGD
    & $40.15 \pm 0.02$
    & $0.973 \pm 0.018$
    & $8.89 \pm 2.73$
    & $0.691 \pm 0.103$
    & $0.081 \pm 0.137$
    & $0.93 \pm 0.37$ \\
    APGD~\cite{croce2020apgd}
    & $42.10 \pm 0.30$
    & $0.980 \pm 0.017$
    & $25.18 \pm 1.23$
    & $0.953 \pm 0.022$
    & $0.618 \pm 0.308$
    & $0.95 \pm 0.36$ \\
    LogExp
    & $43.42 \pm 0.33$
    & $0.984 \pm 0.004$
    & $21.30 \pm 1.52$
    & $0.910 \pm 0.090$
    & $0.488 \pm 0.117$
    & $0.87 \pm 0.35$ \\
    T-MLA
    & $\mathbf{55.82 \pm 0.87}$
    & $\mathbf{0.999 \pm 0.001}$
    & $25.03 \pm 0.33$
    & $0.969 \pm 0.010$
    & $0.728 \pm 0.140$
    & $0.85 \pm 0.37$ \\
    \bottomrule
  \end{tabular}
  \end{adjustbox}}
\end{table*}

\begin{table}[t]
\centering
\caption{Detailed evaluation results for Cheng2020-Anchor model on Kodak dataset}
\label{tab:chenganchor_results_kodak}
\small
\begin{tabular}{@{}c|cc|cccc@{}}
\toprule
\multirow{2}{*}{\begin{tabular}[c]{@{}c@{}}Image\\(kodim)\end{tabular}} & \multicolumn{2}{c|}{\textbf{Stealth} $\uparrow$} & \multicolumn{4}{c}{\textbf{Attack Success} $\downarrow$} \\
\cmidrule(lr){2-3} \cmidrule(lr){4-7}
& PSNR & VIF & PSNR & SSIM & VIF & BPP \\
\midrule
01 & 55.1 & 0.999 & 25.2 & 0.974 & 0.840 & 1.42 \\
02 & 54.1 & 0.998 & 24.9 & 0.954 & 0.500 & 0.70 \\
03 & 55.1 & 0.998 & 24.9 & 0.972 & 0.535 & 0.51 \\
04 & 55.1 & 0.999 & 25.0 & 0.961 & 0.610 & 0.70 \\
05 & 55.0 & 1.000 & 24.7 & 0.977 & 0.875 & 1.34 \\
06 & 55.0 & 0.999 & 25.0 & 0.972 & 0.738 & 1.07 \\
07 & 55.1 & 0.999 & 25.2 & 0.978 & 0.724 & 0.61 \\
08 & 55.0 & 1.000 & 25.0 & 0.972 & 0.911 & 1.44 \\
09 & 55.1 & 0.999 & 24.6 & 0.964 & 0.655 & 0.52 \\
10 & 55.1 & 0.999 & 24.5 & 0.965 & 0.591 & 0.56 \\
11 & 55.0 & 0.999 & 24.9 & 0.968 & 0.744 & 0.93 \\
12 & 55.3 & 0.998 & 24.8 & 0.963 & 0.575 & 0.57 \\
13 & 55.0 & 1.000 & 25.1 & 0.971 & 0.882 & 1.82 \\
14 & 55.0 & 0.999 & 25.0 & 0.968 & 0.795 & 1.17 \\
15 & 55.1 & 0.999 & 24.7 & 0.962 & 0.653 & 0.65 \\
16 & 55.1 & 0.998 & 25.4 & 0.970 & 0.569 & 0.76 \\
17 & 55.1 & 0.999 & 24.8 & 0.967 & 0.623 & 0.64 \\
18 & 55.0 & 0.999 & 25.0 & 0.962 & 0.798 & 1.15 \\
19 & 55.0 & 0.999 & 25.0 & 0.965 & 0.757 & 0.85 \\
20 & 55.0 & 0.999 & 25.2 & 0.971 & 0.814 & 0.59 \\
21 & 55.0 & 0.999 & 24.7 & 0.966 & 0.766 & 0.92 \\
22 & 55.0 & 0.999 & 25.0 & 0.960 & 0.652 & 0.95 \\
23 & 55.0 & 0.999 & 25.3 & 0.969 & 0.622 & 0.44 \\
24 & 55.0 & 0.999 & 24.9 & 0.972 & 0.807 & 1.13 \\
\midrule
\multicolumn{7}{c}{Summary Statistics} \\
\midrule
Mean & 55.0 & 0.999 & 24.9 & 0.968 & 0.710 & 0.89 \\
Std & 0.2 & 0.001 & 0.2 & 0.006 & 0.116 & 0.34 \\
\bottomrule
\end{tabular}
\end{table}

\begin{table}[h!]
    \centering
    \caption{Detailed evaluation results for Cheng2020-Anchor model on CLIC subset}
    \label{tab:chenganchor_results_clic}
    \small
    \begin{tabular}{@{}c|cc|cccc@{}}
    \toprule
    \multirow{2}{*}{\begin{tabular}[c]{@{}c@{}}Image\\(idx)\end{tabular}} & \multicolumn{2}{c|}{\textbf{Stealth} $\uparrow$} & \multicolumn{4}{c}{\textbf{Attack Success} $\downarrow$} \\
    \cmidrule(lr){2-3} \cmidrule(lr){4-7}
    & PSNR & VIF & PSNR & SSIM & VIF & BPP \\
    \midrule
    05 & 57.1 & 0.999 & 24.9 & 0.949 & 0.713 & 0.65 \\
    09 & 57.0 & 0.999 & 24.6 & 0.968 & 0.513 & 0.55 \\
    16 & 57.0 & 0.997 & 25.3 & 0.959 & 0.314 & 0.22 \\
    19 & 57.0 & 0.998 & 25.3 & 0.963 & 0.493 & 0.68 \\
    24 & 57.1 & 0.999 & 25.1 & 0.947 & 0.689 & 0.48 \\
    34 & 57.0 & 1.000 & 25.9 & 0.967 & 0.794 & 1.08 \\
    35 & 57.0 & 0.999 & 24.3 & 0.969 & 0.689 & 0.53 \\
    38 & 57.0 & 1.000 & 25.6 & 0.983 & 0.804 & 0.40 \\
    42 & 57.1 & 0.998 & 24.3 & 0.952 & 0.380 & 0.22 \\
    43 & 57.0 & 0.999 & 24.8 & 0.953 & 0.722 & 0.47 \\
    45 & 57.0 & 0.999 & 25.6 & 0.973 & 0.714 & 0.30 \\
    48 & 55.1 & 0.999 & 24.7 & 0.979 & 0.577 & 0.36 \\
    50 & 57.1 & 1.000 & 25.6 & 0.981 & 0.867 & 0.40 \\
    53 & 57.1 & 1.000 & 24.5 & 0.987 & 0.855 & 0.71 \\
    68 & 57.0 & 0.999 & 24.6 & 0.975 & 0.823 & 0.56 \\
    71 & 57.0 & 0.999 & 24.5 & 0.978 & 0.604 & 0.43 \\
    82 & 57.1 & 0.999 & 24.9 & 0.956 & 0.543 & 0.46 \\
    94 & 57.1 & 0.999 & 24.9 & 0.950 & 0.574 & 0.31 \\
    \bottomrule
    \end{tabular}
    \end{table}

\begin{table}[h!]
    \centering
    \caption{Detailed evaluation results for Cheng2020-Anchor model on CLIC subset (\textbf{continued})}
    \label{tab:chenganchor_results_div2k_stats}
    \small
    \begin{tabular}{@{}c|cc|cccc@{}}
    \toprule
    \multirow{2}{*}{\begin{tabular}[c]{@{}c@{}}Image\\(idx)\end{tabular}} & \multicolumn{2}{c|}{\textbf{Stealth} $\uparrow$} & \multicolumn{4}{c}{\textbf{Attack Success} $\downarrow$} \\
    \cmidrule(lr){2-3} \cmidrule(lr){4-7}
    & PSNR & VIF & PSNR & SSIM & VIF & BPP \\
    \midrule
    96 & 55.3 & 0.999 & 25.3 & 0.928 & 0.818 & 0.78 \\
    97 & 57.1 & 0.999 & 24.7 & 0.967 & 0.648 & 0.48 \\
    99 & 56.7 & 0.999 & 25.6 & 0.939 & 0.665 & 1.04 \\
    100 & 57.0 & 1.000 & 24.9 & 0.981 & 0.913 & 1.19 \\
    105 & 56.9 & 1.000 & 24.5 & 0.974 & 0.831 & 0.45 \\
    108 & 57.1 & 1.000 & 25.1 & 0.967 & 0.785 & 1.34 \\
    109 & 54.4 & 0.999 & 25.1 & 0.974 & 0.703 & 0.41 \\
    110 & 57.0 & 1.000 & 23.9 & 0.981 & 0.774 & 0.45 \\
    116 & 57.0 & 1.000 & 24.8 & 0.976 & 0.821 & 0.55 \\
    118 & 55.9 & 0.999 & 25.3 & 0.969 & 0.796 & 1.16 \\
    122 & 57.0 & 0.997 & 25.2 & 0.969 & 0.254 & 0.26 \\
    127 & 57.0 & 1.000 & 24.8 & 0.955 & 0.893 & 1.02 \\
    128 & 55.0 & 0.997 & 25.0 & 0.981 & 0.397 & 0.23 \\
    130 & 54.2 & 1.000 & 25.6 & 0.985 & 0.875 & 1.23 \\
    134 & 57.2 & 0.999 & 26.2 & 0.981 & 0.604 & 0.28 \\
    139 & 57.1 & 0.998 & 24.8 & 0.975 & 0.380 & 0.25 \\
    141 & 56.9 & 0.999 & 24.3 & 0.960 & 0.655 & 0.32 \\
    144 & 57.1 & 1.000 & 25.0 & 0.980 & 0.780 & 0.61 \\
    145 & 55.4 & 0.998 & 25.3 & 0.939 & 0.609 & 0.71 \\
    147 & 57.0 & 0.999 & 25.1 & 0.984 & 0.652 & 0.38 \\
    155 & 55.0 & 0.999 & 24.6 & 0.980 & 0.665 & 0.37 \\
    160 & 57.1 & 1.000 & 25.7 & 0.986 & 0.888 & 0.69 \\
    161 & 57.1 & 1.000 & 26.1 & 0.973 & 0.779 & 0.36 \\
    167 & 57.1 & 1.000 & 25.1 & 0.974 & 0.914 & 1.11 \\
    168 & 57.1 & 1.000 & 24.6 & 0.982 & 0.719 & 0.27 \\
    172 & 57.0 & 0.995 & 25.5 & 0.975 & 0.287 & 0.36 \\
    174 & 57.0 & 0.998 & 25.1 & 0.958 & 0.374 & 0.37 \\
    175 & 57.1 & 0.999 & 25.0 & 0.963 & 0.714 & 0.47 \\
    176 & 57.1 & 0.999 & 25.0 & 0.942 & 0.705 & 0.72 \\
    177 & 57.1 & 1.000 & 23.7 & 0.977 & 0.793 & 0.43 \\
    189 & 55.1 & 0.996 & 24.7 & 0.965 & 0.326 & 0.32 \\
    199 & 57.1 & 1.000 & 25.0 & 0.980 & 0.882 & 0.80 \\
    203 & 57.1 & 1.000 & 24.5 & 0.979 & 0.877 & 0.83 \\
    204 & 57.0 & 1.000 & 25.1 & 0.962 & 0.921 & 1.06 \\
    209 & 57.0 & 1.000 & 24.4 & 0.963 & 0.722 & 0.36 \\
    212 & 57.1 & 1.000 & 24.9 & 0.981 & 0.767 & 0.34 \\
    214 & 57.0 & 0.999 & 24.6 & 0.979 & 0.523 & 0.28 \\
    219 & 54.7 & 0.999 & 25.2 & 0.978 & 0.735 & 0.71 \\
    221 & 57.1 & 1.000 & 25.6 & 0.959 & 0.817 & 0.77 \\
    227 & 57.1 & 1.000 & 25.1 & 0.974 & 0.838 & 1.04 \\
    231 & 57.0 & 0.998 & 25.0 & 0.960 & 0.396 & 0.37 \\
    232 & 57.2 & 1.000 & 26.0 & 0.955 & 0.825 & 1.68 \\
    235 & 57.0 & 1.000 & 25.0 & 0.969 & 0.876 & 0.94 \\
    242 & 57.1 & 0.997 & 24.9 & 0.945 & 0.344 & 0.26 \\
    246 & 55.2 & 0.999 & 25.9 & 0.986 & 0.749 & 0.29 \\
    \midrule
    \multicolumn{7}{c}{Summary Statistics} \\
    \midrule
    Mean & 56.7 & 0.999 & 25.0 & 0.969 & 0.687 & 0.59 \\
    Std & 0.9 & 0.001 & 0.5 & 0.014 & 0.178 & 0.34 \\
    \bottomrule
    \end{tabular}
    \end{table}

\begin{table}[h!]
    \centering
    \caption{Detailed evaluation results for Cheng2020-Anchor model on DIV2K subset}
    \label{tab:chenganchor_results_div2k}
    \small
    \begin{tabular}{@{}c|cc|cccc@{}}
    \toprule
    \multirow{2}{*}{\begin{tabular}[c]{@{}c@{}}Image\\(idx)\end{tabular}} & \multicolumn{2}{c|}{\textbf{Stealth} $\uparrow$} & \multicolumn{4}{c}{\textbf{Attack Success} $\downarrow$} \\
    \cmidrule(lr){2-3} \cmidrule(lr){4-7}
    & PSNR & VIF & PSNR & SSIM & VIF & BPP \\
    \midrule
    01 & 57.0 & 0.999 & 25.0 & 0.968 & 0.704 & 1.19 \\
    16 & 55.0 & 0.999 & 25.1 & 0.966 & 0.828 & 1.07 \\
    35 & 55.0 & 0.999 & 24.8 & 0.974 & 0.804 & 1.38 \\
    49 & 55.0 & 0.999 & 25.1 & 0.972 & 0.813 & 1.40 \\
    51 & 57.1 & 1.000 & 25.8 & 0.956 & 0.817 & 1.26 \\
    53 & 57.0 & 1.000 & 24.6 & 0.977 & 0.798 & 1.30 \\
    55 & 53.5 & 0.998 & 25.0 & 0.966 & 0.695 & 0.64 \\
    90 & 53.4 & 1.000 & 25.1 & 0.980 & 0.882 & 0.53 \\
    106 & 55.0 & 1.000 & 25.0 & 0.981 & 0.896 & 1.29 \\
    110 & 55.1 & 0.999 & 25.0 & 0.972 & 0.605 & 0.42 \\
    111 & 55.0 & 1.000 & 24.8 & 0.983 & 0.900 & 0.80 \\
    115 & 56.6 & 0.999 & 25.1 & 0.953 & 0.679 & 1.26 \\
    121 & 55.0 & 0.999 & 24.9 & 0.977 & 0.661 & 0.49 \\
    123 & 54.9 & 1.000 & 25.2 & 0.966 & 0.847 & 0.80 \\
    127 & 55.0 & 1.000 & 25.1 & 0.984 & 0.902 & 0.80 \\
    130 & 57.1 & 0.999 & 25.1 & 0.977 & 0.474 & 0.41 \\
    138 & 55.0 & 0.995 & 25.1 & 0.986 & 0.311 & 0.27 \\
    139 & 56.6 & 1.000 & 25.3 & 0.942 & 0.896 & 1.76 \\
    141 & 53.7 & 0.999 & 25.1 & 0.969 & 0.869 & 1.42 \\
    155 & 52.9 & 0.999 & 25.2 & 0.982 & 0.963 & 0.78 \\
    201 & 57.0 & 1.000 & 25.8 & 0.967 & 0.841 & 1.33 \\
    213 & 57.0 & 1.000 & 25.1 & 0.963 & 0.971 & 2.47 \\
    229 & 53.9 & 0.998 & 25.1 & 0.976 & 0.633 & 0.57 \\
    268 & 57.1 & 0.999 & 24.6 & 0.968 & 0.528 & 0.47 \\
    269 & 57.1 & 0.999 & 25.4 & 0.957 & 0.699 & 0.58 \\
    297 & 55.3 & 1.000 & 24.9 & 0.978 & 0.896 & 1.99 \\
    299 & 53.2 & 1.000 & 25.4 & 0.977 & 0.907 & 1.46 \\
    305 & 53.1 & 0.999 & 25.1 & 0.966 & 0.786 & 1.52 \\
    324 & 57.0 & 1.000 & 25.0 & 0.941 & 0.785 & 1.50 \\
    336 & 57.0 & 1.000 & 24.7 & 0.965 & 0.682 & 0.95 \\
    349 & 52.6 & 0.999 & 24.8 & 0.976 & 0.910 & 1.30 \\
    351 & 57.1 & 1.000 & 24.5 & 0.976 & 0.893 & 1.03 \\
    363 & 57.1 & 1.000 & 24.9 & 0.962 & 0.868 & 1.79 \\
    365 & 56.9 & 1.000 & 24.0 & 0.973 & 0.702 & 0.74 \\
    374 & 54.0 & 0.999 & 25.4 & 0.970 & 0.769 & 1.09 \\
    397 & 52.9 & 0.998 & 25.3 & 0.970 & 0.763 & 1.08 \\
    404 & 57.1 & 1.000 & 24.6 & 0.953 & 0.806 & 1.06 \\
    407 & 57.0 & 1.000 & 24.5 & 0.961 & 0.675 & 0.89 \\
    412 & 55.8 & 1.000 & 25.1 & 0.974 & 0.916 & 1.32 \\
    468 & 57.0 & 1.000 & 25.3 & 0.969 & 0.840 & 0.86 \\
    502 & 57.1 & 1.000 & 25.0 & 0.980 & 0.873 & 1.44 \\
    507 & 57.0 & 0.999 & 24.8 & 0.975 & 0.474 & 0.34 \\
    524 & 57.0 & 1.000 & 25.0 & 0.974 & 0.834 & 0.73 \\
    525 & 57.2 & 0.999 & 25.1 & 0.941 & 0.636 & 0.66 \\
    533 & 57.0 & 1.000 & 24.7 & 0.977 & 0.899 & 0.96 \\
    539 & 57.0 & 1.000 & 24.8 & 0.979 & 0.870 & 0.67 \\
    578 & 57.1 & 0.999 & 24.7 & 0.965 & 0.628 & 0.55 \\
    584 & 55.0 & 1.000 & 25.1 & 0.969 & 0.848 & 1.20 \\
    592 & 57.1 & 1.000 & 24.8 & 0.970 & 0.785 & 1.09 \\
    606 & 55.1 & 1.000 & 25.5 & 0.977 & 0.876 & 1.17 \\
    618 & 57.1 & 1.000 & 24.4 & 0.973 & 0.722 & 0.53 \\
    627 & 57.0 & 1.000 & 24.3 & 0.976 & 0.852 & 1.24 \\
    632 & 57.0 & 0.999 & 25.0 & 0.954 & 0.539 & 1.14 \\
    644 & 57.0 & 1.000 & 25.5 & 0.975 & 0.944 & 1.21 \\
    \bottomrule
    \end{tabular}
\end{table}

\begin{table}[h!]
    \centering
    \caption{Detailed evaluation results for Cheng2020-Anchor model on DIV2K subset (\textbf{continued})}
    \label{tab:chenganchor_results_div2k_stats}
    \small
    \begin{tabular}{@{}c|cc|cccc@{}}
    \toprule
    \multirow{2}{*}{\begin{tabular}[c]{@{}c@{}}Image\\(idx)\end{tabular}} & \multicolumn{2}{c|}{\textbf{Stealth} $\uparrow$} & \multicolumn{4}{c}{\textbf{Attack Success} $\downarrow$} \\
    \cmidrule(lr){2-3} \cmidrule(lr){4-7}
    & PSNR & VIF & PSNR & SSIM & VIF & BPP \\
    \midrule
    657 & 55.0 & 1.000 & 24.9 & 0.977 & 0.910 & 1.32 \\
    677 & 57.1 & 1.000 & 25.2 & 0.978 & 0.843 & 0.87 \\
    681 & 57.1 & 1.000 & 24.3 & 0.980 & 0.888 & 1.01 \\
    683 & 54.8 & 0.999 & 25.1 & 0.964 & 0.722 & 0.91 \\
    727 & 57.0 & 0.999 & 25.1 & 0.978 & 0.649 & 0.34 \\
    740 & 55.1 & 1.000 & 24.6 & 0.970 & 0.858 & 1.60 \\
    749 & 57.1 & 0.999 & 24.6 & 0.968 & 0.732 & 0.58 \\
    761 & 55.1 & 1.000 & 25.6 & 0.965 & 0.872 & 1.44 \\
    784 & 53.3 & 0.999 & 25.2 & 0.955 & 0.766 & 0.94 \\
    788 & 54.2 & 0.999 & 25.0 & 0.966 & 0.730 & 1.67 \\
    797 & 52.7 & 0.999 & 25.2 & 0.937 & 0.856 & 1.82 \\
    \midrule
    \multicolumn{7}{c}{Summary Statistics} \\
    \midrule
    Mean & 55.8 & 0.999 & 25.0 & 0.970 & 0.787 & 1.08 \\
    Std & 1.5 & 0.001 & 0.3 & 0.011 & 0.124 & 0.43 \\
    \bottomrule
    \end{tabular}
    \end{table}

\begin{table}[h!]
\centering
\caption{Detailed evaluation results for Cheng2020-Attention model on Kodak dataset}
\label{tab:chengattn_results_kodak}
\small
\begin{tabular}{@{}c|cc|cccc@{}}
\toprule
\multirow{2}{*}{\begin{tabular}[c]{@{}c@{}}Image\\(kodim)\end{tabular}} & \multicolumn{2}{c|}{\textbf{Stealth} $\uparrow$} & \multicolumn{4}{c}{\textbf{Attack Success} $\downarrow$} \\
\cmidrule(lr){2-3} \cmidrule(lr){4-7}
& PSNR & VIF & PSNR & SSIM & VIF & BPP \\
\midrule
01 & 50.1 & 0.999 & 25.1 & 0.973 & 0.845 & 1.42 \\
02 & 50.0 & 0.995 & 24.8 & 0.955 & 0.576 & 0.71 \\
03 & 49.2 & 0.995 & 25.1 & 0.975 & 0.740 & 0.52 \\
04 & 48.4 & 0.994 & 25.1 & 0.962 & 0.799 & 0.71 \\
05 & 48.4 & 0.999 & 25.1 & 0.980 & 0.960 & 1.36 \\
06 & 48.9 & 0.997 & 25.2 & 0.973 & 0.898 & 1.08 \\
07 & 50.0 & 0.998 & 25.0 & 0.977 & 0.744 & 0.62 \\
08 & 49.6 & 0.999 & 25.1 & 0.973 & 0.927 & 1.45 \\
09 & 49.9 & 0.997 & 25.0 & 0.966 & 0.719 & 0.53 \\
10 & 50.0 & 0.997 & 24.9 & 0.966 & 0.674 & 0.57 \\
11 & 49.1 & 0.997 & 25.1 & 0.969 & 0.896 & 0.94 \\
12 & 48.3 & 0.996 & 25.1 & 0.965 & 0.704 & 0.58 \\
13 & 48.8 & 0.998 & 25.3 & 0.972 & 0.957 & 1.80 \\
14 & 48.7 & 0.997 & 25.1 & 0.969 & 0.917 & 1.19 \\
15 & 48.7 & 0.997 & 25.2 & 0.965 & 0.876 & 0.66 \\
16 & 49.4 & 0.995 & 24.9 & 0.969 & 0.547 & 0.76 \\
17 & 50.0 & 0.997 & 25.0 & 0.969 & 0.729 & 0.65 \\
18 & 49.4 & 0.998 & 25.2 & 0.964 & 0.930 & 1.16 \\
19 & 50.1 & 0.998 & 25.1 & 0.966 & 0.831 & 0.86 \\
20 & 49.5 & 0.997 & 25.2 & 0.971 & 0.828 & 0.60 \\
21 & 48.7 & 0.998 & 25.2 & 0.968 & 0.895 & 0.92 \\
22 & 48.6 & 0.996 & 25.1 & 0.962 & 0.820 & 0.95 \\
23 & 48.7 & 0.995 & 24.9 & 0.965 & 0.660 & 0.45 \\
24 & 50.1 & 0.998 & 25.0 & 0.973 & 0.844 & 1.13 \\
\midrule
\multicolumn{7}{c}{Summary Statistics} \\
\midrule
Mean & 49.2 & 0.997 & 25.1 & 0.969 & 0.813 & 0.89 \\
Std & 0.7 & 0.001 & 0.1 & 0.005 & 0.111 & 0.34 \\
\bottomrule
\end{tabular}
\end{table}

\begin{table}[h!]
\centering
\caption{Detailed evaluation results for Cheng2020-Attention model on CLIC subset}
\label{tab:chengattn_results_clic}
\small
\begin{tabular}{@{}c|cc|cccc@{}}
\toprule
\multirow{2}{*}{\begin{tabular}[c]{@{}c@{}}Image\\(idx)\end{tabular}} & \multicolumn{2}{c|}{\textbf{Stealth} $\uparrow$} & \multicolumn{4}{c}{\textbf{Attack Success} $\downarrow$} \\
\cmidrule(lr){2-3} \cmidrule(lr){4-7}
& PSNR & VIF & PSNR & SSIM & VIF & BPP \\
\midrule
05 & 48.2 & 0.996 & 25.3 & 0.950 & 0.819 & 0.65 \\
09 & 48.1 & 0.993 & 25.2 & 0.969 & 0.681 & 0.57 \\
16 & 48.0 & 0.982 & 24.9 & 0.960 & 0.581 & 0.24 \\
19 & 48.9 & 0.988 & 25.0 & 0.962 & 0.573 & 0.70 \\
24 & 49.5 & 0.997 & 25.3 & 0.946 & 0.868 & 0.49 \\
34 & 48.0 & 0.996 & 25.1 & 0.966 & 0.893 & 1.08 \\
35 & 50.1 & 0.997 & 25.0 & 0.970 & 0.769 & 0.53 \\
38 & 48.4 & 0.997 & 25.1 & 0.985 & 0.922 & 0.41 \\
42 & 48.1 & 0.990 & 25.2 & 0.955 & 0.810 & 0.23 \\
43 & 49.1 & 0.997 & 25.0 & 0.953 & 0.792 & 0.49 \\
45 & 49.3 & 0.997 & 25.1 & 0.974 & 0.886 & 0.32 \\
48 & 49.6 & 0.996 & 25.0 & 0.979 & 0.605 & 0.38 \\
50 & 49.1 & 0.999 & 25.0 & 0.982 & 0.951 & 0.41 \\
53 & 48.7 & 0.999 & 25.2 & 0.988 & 0.930 & 0.73 \\
68 & 49.0 & 0.998 & 25.1 & 0.976 & 0.941 & 0.57 \\
71 & 48.9 & 0.996 & 25.2 & 0.982 & 0.813 & 0.44 \\
82 & 49.2 & 0.994 & 25.2 & 0.958 & 0.760 & 0.48 \\
94 & 48.4 & 0.995 & 25.0 & 0.953 & 0.854 & 0.32 \\
96 & 49.5 & 0.998 & 25.2 & 0.929 & 0.923 & 0.79 \\
97 & 48.8 & 0.995 & 25.4 & 0.969 & 0.870 & 0.49 \\
99 & 48.7 & 0.995 & 25.4 & 0.939 & 0.834 & 1.03 \\
100 & 49.3 & 0.999 & 25.4 & 0.981 & 0.930 & 1.20 \\
105 & 48.6 & 0.998 & 25.3 & 0.975 & 0.928 & 0.46 \\
108 & 49.5 & 0.997 & 25.1 & 0.968 & 0.913 & 1.34 \\
109 & 50.2 & 0.997 & 25.1 & 0.974 & 0.756 & 0.41 \\
110 & 48.6 & 0.998 & 25.2 & 0.985 & 0.931 & 0.47 \\
116 & 49.2 & 0.998 & 25.2 & 0.978 & 0.938 & 0.57 \\
118 & 48.7 & 0.998 & 25.0 & 0.971 & 0.918 & 1.17 \\
122 & 50.1 & 0.983 & 25.2 & 0.970 & 0.319 & 0.27 \\
127 & 49.4 & 0.999 & 25.1 & 0.957 & 0.965 & 1.03 \\
128 & 48.8 & 0.991 & 25.3 & 0.984 & 0.745 & 0.24 \\
130 & 48.1 & 0.998 & 25.2 & 0.985 & 0.951 & 1.23 \\
134 & 47.5 & 0.990 & 25.2 & 0.982 & 0.950 & 0.29 \\
139 & 50.0 & 0.993 & 25.0 & 0.977 & 0.569 & 0.26 \\
141 & 48.3 & 0.995 & 25.2 & 0.964 & 0.871 & 0.33 \\
144 & 49.2 & 0.998 & 25.0 & 0.982 & 0.867 & 0.63 \\
145 & 49.2 & 0.995 & 24.9 & 0.942 & 0.804 & 0.73 \\
147 & 47.2 & 0.992 & 25.1 & 0.983 & 0.823 & 0.40 \\
155 & 49.6 & 0.997 & 25.0 & 0.983 & 0.868 & 0.38 \\
160 & 48.3 & 0.999 & 25.4 & 0.986 & 0.932 & 0.73 \\
161 & 50.1 & 0.997 & 25.1 & 0.972 & 0.792 & 0.38 \\
167 & 49.3 & 0.999 & 25.3 & 0.974 & 0.964 & 1.12 \\
168 & 50.2 & 0.998 & 24.9 & 0.984 & 0.791 & 0.28 \\
172 & 47.4 & 0.973 & 25.4 & 0.976 & 0.668 & 0.39 \\
174 & 48.1 & 0.988 & 25.5 & 0.957 & 0.420 & 0.39 \\
175 & 48.8 & 0.996 & 25.0 & 0.964 & 0.845 & 0.49 \\
176 & 50.2 & 0.997 & 25.0 & 0.942 & 0.735 & 0.73 \\
177 & 48.5 & 0.998 & 25.7 & 0.979 & 0.872 & 0.45 \\
189 & 49.5 & 0.987 & 24.9 & 0.967 & 0.647 & 0.33 \\
199 & 49.5 & 0.999 & 25.2 & 0.982 & 0.957 & 0.82 \\
203 & 48.8 & 0.999 & 25.2 & 0.981 & 0.959 & 0.83 \\
204 & 50.1 & 0.999 & 25.0 & 0.954 & 0.922 & 1.10 \\
209 & 48.1 & 0.997 & 25.0 & 0.965 & 0.836 & 0.37 \\
212 & 49.1 & 0.998 & 25.5 & 0.982 & 0.875 & 0.35 \\
214 & 49.3 & 0.996 & 25.0 & 0.982 & 0.828 & 0.29 \\
219 & 47.6 & 0.996 & 25.8 & 0.980 & 0.953 & 0.72 \\
\bottomrule
\end{tabular}
\end{table}

\begin{table}[h!]
\centering
\caption{Detailed evaluation results for Cheng2020-Attention model on CLIC subset (\textbf{continued})}
\label{tab:chengattn_results_clic}
\small
\begin{tabular}{@{}c|cc|cccc@{}}
\toprule
\multirow{2}{*}{\begin{tabular}[c]{@{}c@{}}Image\\(idx)\end{tabular}} & \multicolumn{2}{c|}{\textbf{Stealth} $\uparrow$} & \multicolumn{4}{c}{\textbf{Attack Success} $\downarrow$} \\
\cmidrule(lr){2-3} \cmidrule(lr){4-7}
& PSNR & VIF & PSNR & SSIM & VIF & BPP \\
\midrule
221 & 48.2 & 0.998 & 25.5 & 0.959 & 0.854 & 0.79 \\
227 & 49.5 & 0.999 & 25.1 & 0.977 & 0.957 & 1.05 \\
231 & 48.2 & 0.990 & 25.0 & 0.962 & 0.721 & 0.39 \\
232 & 48.3 & 0.996 & 25.1 & 0.954 & 0.908 & 1.66 \\
235 & 49.8 & 0.999 & 25.2 & 0.968 & 0.945 & 0.94 \\
242 & 48.9 & 0.986 & 25.0 & 0.947 & 0.588 & 0.27 \\
246 & 48.8 & 0.995 & 25.2 & 0.986 & 0.859 & 0.31 \\
\midrule
\multicolumn{7}{c}{Summary Statistics} \\
\midrule
Mean & 48.9 & 0.995 & 25.2 & 0.970 & 0.831 & 0.61 \\
Std & 0.7 & 0.005 & 0.2 & 0.013 & 0.133 & 0.33 \\
\bottomrule
\end{tabular}
\end{table}

\begin{table}[h!]
\centering
\caption{Detailed evaluation results for Cheng2020-Attention model on DIV2K subset}
\label{tab:chengattn_results_div2k}
\small
\begin{tabular}{@{}c|cc|cccc@{}}
\toprule
\multirow{2}{*}{\begin{tabular}[c]{@{}c@{}}Image\\(idx)\end{tabular}} & \multicolumn{2}{c|}{\textbf{Stealth} $\uparrow$} & \multicolumn{4}{c}{\textbf{Attack Success} $\downarrow$} \\
\cmidrule(lr){2-3} \cmidrule(lr){4-7}
& PSNR & VIF & PSNR & SSIM & VIF & BPP \\
\midrule
01 & 50.0 & 0.997 & 24.9 & 0.967 & 0.816 & 1.20 \\
16 & 49.2 & 0.998 & 25.0 & 0.967 & 0.928 & 1.09 \\
35 & 49.2 & 0.998 & 25.3 & 0.975 & 0.834 & 1.37 \\
49 & 48.7 & 0.998 & 25.3 & 0.973 & 0.912 & 1.41 \\
51 & 49.4 & 0.998 & 25.1 & 0.957 & 0.906 & 1.26 \\
53 & 48.7 & 0.998 & 25.1 & 0.979 & 0.910 & 1.32 \\
55 & 49.2 & 0.995 & 25.1 & 0.967 & 0.860 & 0.65 \\
90 & 49.9 & 0.999 & 25.1 & 0.978 & 0.850 & 0.54 \\
106 & 49.9 & 0.999 & 25.0 & 0.980 & 0.903 & 1.29 \\
110 & 50.0 & 0.996 & 25.4 & 0.975 & 0.800 & 0.43 \\
111 & 47.9 & 0.998 & 25.0 & 0.983 & 0.966 & 0.85 \\
115 & 49.7 & 0.997 & 25.1 & 0.954 & 0.824 & 1.26 \\
121 & 48.1 & 0.996 & 25.5 & 0.982 & 0.940 & 0.51 \\
123 & 49.0 & 0.999 & 25.1 & 0.967 & 0.930 & 0.82 \\
127 & 48.5 & 0.996 & 26.0 & 0.975 & 0.873 & 0.61 \\
130 & 49.5 & 0.992 & 25.2 & 0.980 & 0.725 & 0.42 \\
138 & 48.4 & 0.981 & 24.9 & 0.988 & 0.666 & 0.28 \\
139 & 50.0 & 0.999 & 25.0 & 0.942 & 0.878 & 1.76 \\
141 & 50.0 & 0.999 & 25.0 & 0.968 & 0.871 & 1.43 \\
155 & 47.7 & 0.998 & 25.8 & 0.982 & 0.949 & 0.79 \\
201 & 48.7 & 0.998 & 25.3 & 0.968 & 0.931 & 1.35 \\
213 & 50.1 & 1.000 & 25.0 & 0.964 & 0.978 & 2.45 \\
229 & 48.6 & 0.994 & 24.9 & 0.976 & 0.701 & 0.57 \\
268 & 48.6 & 0.992 & 25.2 & 0.970 & 0.797 & 0.48 \\
269 & 48.4 & 0.996 & 25.0 & 0.958 & 0.875 & 0.60 \\
297 & 50.0 & 0.999 & 24.9 & 0.977 & 0.913 & 1.98 \\
299 & 48.7 & 0.999 & 25.1 & 0.977 & 0.959 & 1.47 \\
305 & 48.6 & 0.998 & 25.0 & 0.966 & 0.918 & 1.52 \\
324 & 50.1 & 0.998 & 24.9 & 0.938 & 0.801 & 1.51 \\
336 & 50.0 & 0.997 & 25.6 & 0.969 & 0.858 & 0.95 \\
349 & 49.3 & 0.999 & 25.2 & 0.975 & 0.908 & 1.30 \\
351 & 49.9 & 0.999 & 24.9 & 0.975 & 0.899 & 1.05 \\
363 & 50.0 & 0.999 & 25.7 & 0.963 & 0.946 & 1.77 \\
365 & 48.8 & 0.997 & 25.1 & 0.977 & 0.918 & 0.74 \\
374 & 50.1 & 0.998 & 25.1 & 0.969 & 0.770 & 1.10 \\
397 & 50.0 & 0.997 & 24.6 & 0.969 & 0.777 & 1.09 \\
\bottomrule
\end{tabular}
\end{table}

\begin{table}[h!]
\centering
\caption{Detailed evaluation results for Cheng2020-Attention model on DIV2K subset (\textbf{continued})}
\label{tab:chengattn_results_div2k_stats}
\small
\begin{tabular}{@{}c|cc|cccc@{}}
\toprule
\multirow{2}{*}{\begin{tabular}[c]{@{}c@{}}Image\\(idx)\end{tabular}} & \multicolumn{2}{c|}{\textbf{Stealth} $\uparrow$} & \multicolumn{4}{c}{\textbf{Attack Success} $\downarrow$} \\
\cmidrule(lr){2-3} \cmidrule(lr){4-7}
& PSNR & VIF & PSNR & SSIM & VIF & BPP \\
\midrule
404 & 49.3 & 0.998 & 25.0 & 0.954 & 0.826 & 1.09 \\
407 & 49.3 & 0.997 & 25.3 & 0.963 & 0.885 & 0.90 \\
412 & 50.0 & 0.999 & 25.0 & 0.974 & 0.919 & 1.34 \\
468 & 48.3 & 0.998 & 25.1 & 0.968 & 0.931 & 0.87 \\
502 & 48.3 & 0.999 & 25.5 & 0.982 & 0.958 & 1.44 \\
507 & 48.2 & 0.992 & 25.8 & 0.978 & 0.757 & 0.35 \\
524 & 48.5 & 0.998 & 25.3 & 0.975 & 0.902 & 0.74 \\
525 & 48.2 & 0.996 & 25.2 & 0.938 & 0.704 & 0.68 \\
533 & 50.0 & 0.999 & 25.1 & 0.977 & 0.896 & 0.97 \\
539 & 50.1 & 0.999 & 24.8 & 0.979 & 0.903 & 0.69 \\
578 & 48.8 & 0.995 & 25.1 & 0.967 & 0.866 & 0.56 \\
584 & 50.0 & 0.999 & 24.9 & 0.968 & 0.867 & 1.20 \\
592 & 48.7 & 0.998 & 25.0 & 0.973 & 0.937 & 1.10 \\
606 & 48.5 & 0.999 & 25.1 & 0.978 & 0.957 & 1.20 \\
618 & 50.0 & 0.998 & 24.8 & 0.969 & 0.747 & 0.54 \\
627 & 49.5 & 0.999 & 25.7 & 0.978 & 0.963 & 1.25 \\
632 & 48.4 & 0.992 & 25.4 & 0.954 & 0.721 & 1.17 \\
644 & 48.8 & 0.999 & 25.0 & 0.976 & 0.966 & 1.22 \\
657 & 48.7 & 0.999 & 25.4 & 0.978 & 0.971 & 1.34 \\
677 & 50.0 & 0.999 & 25.1 & 0.979 & 0.882 & 0.90 \\
681 & 48.7 & 0.999 & 25.1 & 0.983 & 0.969 & 1.03 \\
683 & 49.5 & 0.997 & 25.3 & 0.965 & 0.799 & 0.93 \\
727 & 48.4 & 0.996 & 25.0 & 0.979 & 0.851 & 0.36 \\
740 & 50.0 & 0.999 & 25.2 & 0.971 & 0.887 & 1.59 \\
749 & 50.1 & 0.999 & 25.3 & 0.969 & 0.834 & 0.59 \\
761 & 48.8 & 0.998 & 25.1 & 0.965 & 0.939 & 1.45 \\
784 & 48.5 & 0.997 & 25.2 & 0.954 & 0.902 & 0.96 \\
788 & 49.2 & 0.997 & 25.8 & 0.968 & 0.907 & 1.68 \\
797 & 50.0 & 0.997 & 24.9 & 0.967 & 0.780 & 1.20 \\
\midrule
\multicolumn{7}{c}{Summary Statistics} \\
\midrule
Mean & 49.2 & 0.997 & 25.2 & 0.971 & 0.873 & 1.08 \\
Std & 0.7 & 0.003 & 0.3 & 0.010 & 0.078 & 0.43 \\
\bottomrule
\end{tabular}
\end{table}

\begin{table}[h!]
\centering
\caption{Detailed evaluation results for LIC-TCM model on Kodak dataset}
\label{tab:tcm_results_kodak}
\small
\begin{tabular}{@{}c|cc|cccc@{}}
\toprule
\multirow{2}{*}{\begin{tabular}[c]{@{}c@{}}Image\\(kodim)\end{tabular}} & \multicolumn{2}{c|}{\textbf{Stealth} $\uparrow$} & \multicolumn{4}{c}{\textbf{Attack Success} $\downarrow$} \\
\cmidrule(lr){2-3} \cmidrule(lr){4-7}
& PSNR & VIF & PSNR & SSIM & VIF & BPP \\
\midrule
01 & 54.9 & 1.000 & 25.0 & 0.975 & 0.811 & 1.63 \\
02 & 54.6 & 0.997 & 26.9 & 0.958 & 0.643 & 0.81 \\
03 & 55.1 & 0.999 & 25.1 & 0.972 & 0.545 & 0.58 \\
04 & 54.9 & 0.999 & 25.6 & 0.953 & 0.618 & 1.26 \\
05 & 55.2 & 1.000 & 25.0 & 0.979 & 0.890 & 1.56 \\
06 & 55.0 & 1.000 & 26.8 & 0.965 & 0.752 & 1.57 \\
07 & 55.0 & 1.000 & 25.0 & 0.962 & 0.703 & 1.36 \\
08 & 55.0 & 1.000 & 27.5 & 0.966 & 0.933 & 1.98 \\
09 & 55.0 & 0.999 & 26.9 & 0.957 & 0.735 & 0.93 \\
10 & 55.2 & 0.999 & 25.1 & 0.963 & 0.614 & 0.64 \\
11 & 55.1 & 0.999 & 25.0 & 0.968 & 0.813 & 1.15 \\
12 & 55.0 & 0.999 & 26.2 & 0.958 & 0.605 & 1.02 \\
13 & 55.1 & 1.000 & 26.8 & 0.971 & 0.886 & 2.48 \\
14 & 55.1 & 1.000 & 25.0 & 0.947 & 0.722 & 2.09 \\
\bottomrule
\end{tabular}
\end{table}

\begin{table}[h!]
    \centering
    \caption{Detailed evaluation results for LIC-TCM model on Kodak dataset (\textbf{continued})}
    \label{tab:tcm_results_kodak}
    \small
    \begin{tabular}{@{}c|cc|cccc@{}}
    \toprule
    \multirow{2}{*}{\begin{tabular}[c]{@{}c@{}}Image\\(kodim)\end{tabular}} & \multicolumn{2}{c|}{\textbf{Stealth} $\uparrow$} & \multicolumn{4}{c}{\textbf{Attack Success} $\downarrow$} \\
    \cmidrule(lr){2-3} \cmidrule(lr){4-7}
    & PSNR & VIF & PSNR & SSIM & VIF & BPP \\
    \midrule
    15 & 55.4 & 0.999 & 24.1 & 0.965 & 0.703 & 0.74 \\
    16 & 55.1 & 0.999 & 24.9 & 0.969 & 0.573 & 0.82 \\
    17 & 55.2 & 0.999 & 25.1 & 0.967 & 0.699 & 0.88 \\
    18 & 54.7 & 1.000 & 26.0 & 0.954 & 0.784 & 1.80 \\
    19 & 55.1 & 1.000 & 25.0 & 0.965 & 0.814 & 1.17 \\
    20 & 54.9 & 1.000 & 25.5 & 0.951 & 0.701 & 1.27 \\
    21 & 55.0 & 1.000 & 27.2 & 0.962 & 0.803 & 1.37 \\
    22 & 55.1 & 1.000 & 25.0 & 0.954 & 0.633 & 1.44 \\
    23 & 55.2 & 0.999 & 25.2 & 0.957 & 0.641 & 1.02 \\
    24 & 55.2 & 1.000 & 25.0 & 0.970 & 0.798 & 1.56 \\
    \midrule
    \multicolumn{7}{c}{Summary Statistics} \\
    \midrule
    Mean & 55.0 & 0.999 & 25.6 & 0.963 & 0.726 & 1.29 \\
    Std & 0.2 & 0.001 & 0.9 & 0.008 & 0.103 & 0.45 \\
    \bottomrule
    \end{tabular}
    \end{table}

\begin{table}[h!]
\centering
\caption{Detailed evaluation results for LIC-TCM model on CLIC subset}
\label{tab:tcm_results_clic}
\small
\begin{tabular}{@{}c|cc|cccc@{}}
\toprule
\multirow{2}{*}{\begin{tabular}[c]{@{}c@{}}Image\\(idx)\end{tabular}} & \multicolumn{2}{c|}{\textbf{Stealth} $\uparrow$} & \multicolumn{4}{c}{\textbf{Attack Success} $\downarrow$} \\
\cmidrule(lr){2-3} \cmidrule(lr){4-7}
& PSNR & VIF & PSNR & SSIM & VIF & BPP \\
\midrule
05 & 53.9 & 0.999 & 25.8 & 0.948 & 0.715 & 1.42 \\
09 & 55.1 & 0.999 & 25.0 & 0.964 & 0.526 & 0.87 \\
16 & 54.5 & 0.996 & 26.4 & 0.949 & 0.378 & 0.68 \\
19 & 54.0 & 0.997 & 25.9 & 0.945 & 0.467 & 1.51 \\
24 & 54.1 & 0.999 & 26.0 & 0.939 & 0.687 & 1.11 \\
34 & 55.1 & 1.000 & 24.8 & 0.968 & 0.738 & 1.32 \\
35 & 54.4 & 0.999 & 26.5 & 0.963 & 0.768 & 0.96 \\
38 & 52.2 & 0.999 & 25.4 & 0.967 & 0.771 & 1.04 \\
42 & 54.5 & 0.998 & 25.6 & 0.937 & 0.402 & 0.84 \\
43 & 53.0 & 0.999 & 25.8 & 0.948 & 0.711 & 0.81 \\
45 & 55.1 & 0.999 & 24.6 & 0.970 & 0.696 & 0.36 \\
48 & 53.8 & 0.999 & 26.8 & 0.972 & 0.670 & 0.75 \\
50 & 54.0 & 1.000 & 26.7 & 0.972 & 0.874 & 0.74 \\
53 & 52.2 & 1.000 & 25.7 & 0.973 & 0.859 & 1.38 \\
68 & 54.4 & 0.999 & 26.0 & 0.971 & 0.835 & 0.72 \\
71 & 55.1 & 0.999 & 24.8 & 0.975 & 0.618 & 0.48 \\
82 & 55.0 & 0.998 & 26.7 & 0.940 & 0.593 & 1.13 \\
94 & 55.0 & 0.999 & 26.1 & 0.939 & 0.652 & 0.87 \\
96 & 51.7 & 0.999 & 26.3 & 0.921 & 0.794 & 1.33 \\
97 & 54.1 & 0.999 & 24.7 & 0.965 & 0.619 & 0.58 \\
99 & 55.0 & 0.999 & 27.2 & 0.940 & 0.685 & 1.62 \\
100 & 54.9 & 1.000 & 25.4 & 0.973 & 0.891 & 1.72 \\
105 & 54.2 & 1.000 & 25.8 & 0.964 & 0.830 & 1.01 \\
108 & 54.2 & 1.000 & 26.4 & 0.950 & 0.738 & 2.26 \\
109 & 55.0 & 0.999 & 26.2 & 0.958 & 0.685 & 1.04 \\
110 & 54.9 & 1.000 & 25.5 & 0.969 & 0.794 & 1.13 \\
116 & 55.3 & 1.000 & 25.1 & 0.972 & 0.836 & 0.81 \\
118 & 51.5 & 0.999 & 25.9 & 0.954 & 0.758 & 1.89 \\
122 & 54.8 & 0.996 & 25.7 & 0.952 & 0.298 & 0.94 \\
127 & 52.3 & 1.000 & 25.6 & 0.948 & 0.893 & 1.60 \\
128 & 54.1 & 0.997 & 25.6 & 0.959 & 0.386 & 0.94 \\
130 & 53.5 & 1.000 & 25.8 & 0.970 & 0.859 & 1.95 \\
134 & 53.5 & 0.998 & 26.4 & 0.969 & 0.553 & 0.73 \\
139 & 55.1 & 0.998 & 25.0 & 0.976 & 0.499 & 0.37 \\
\bottomrule
\end{tabular}
\end{table}

\clearpage

\begin{table}[h!]
\centering
\caption{Detailed evaluation results for LIC-TCM model on CLIC subset (\textbf{continued})}
\label{tab:tcm_results_clic}
\small
\begin{tabular}{@{}c|cc|cccc@{}}
\toprule
\multirow{2}{*}{\begin{tabular}[c]{@{}c@{}}Image\\(idx)\end{tabular}} & \multicolumn{2}{c|}{\textbf{Stealth} $\uparrow$} & \multicolumn{4}{c}{\textbf{Attack Success} $\downarrow$} \\
\cmidrule(lr){2-3} \cmidrule(lr){4-7}
& PSNR & VIF & PSNR & SSIM & VIF & BPP \\
\midrule
141 & 55.0 & 0.999 & 25.3 & 0.953 & 0.650 & 0.86 \\
144 & 52.2 & 0.999 & 26.9 & 0.975 & 0.836 & 1.03 \\
145 & 53.9 & 0.999 & 26.4 & 0.934 & 0.653 & 1.27 \\
147 & 52.3 & 0.998 & 25.5 & 0.969 & 0.592 & 1.05 \\
155 & 53.3 & 0.999 & 25.7 & 0.965 & 0.694 & 0.95 \\
160 & 55.2 & 1.000 & 25.0 & 0.981 & 0.866 & 1.03 \\
161 & 53.6 & 0.999 & 25.9 & 0.960 & 0.773 & 0.81 \\
167 & 51.4 & 1.000 & 25.6 & 0.970 & 0.909 & 1.43 \\
168 & 55.1 & 1.000 & 25.0 & 0.981 & 0.748 & 0.42 \\
172 & 54.9 & 0.995 & 25.9 & 0.961 & 0.238 & 1.00 \\
174 & 55.3 & 0.998 & 25.4 & 0.958 & 0.400 & 0.37 \\
175 & 53.7 & 0.999 & 25.8 & 0.952 & 0.679 & 0.88 \\
176 & 54.9 & 0.999 & 24.8 & 0.941 & 0.663 & 0.76 \\
177 & 52.4 & 0.999 & 26.3 & 0.965 & 0.829 & 0.93 \\
189 & 55.1 & 0.997 & 26.8 & 0.961 & 0.395 & 0.66 \\
199 & 54.9 & 1.000 & 27.3 & 0.977 & 0.907 & 1.06 \\
203 & 52.8 & 1.000 & 26.1 & 0.969 & 0.888 & 1.30 \\
204 & 52.4 & 1.000 & 24.9 & 0.964 & 0.910 & 1.34 \\
209 & 54.1 & 0.999 & 25.2 & 0.949 & 0.731 & 1.00 \\
212 & 53.3 & 0.999 & 26.2 & 0.969 & 0.795 & 0.82 \\
214 & 54.5 & 0.998 & 25.7 & 0.966 & 0.569 & 0.97 \\
219 & 55.0 & 1.000 & 25.6 & 0.958 & 0.714 & 1.48 \\
221 & 51.2 & 0.999 & 25.5 & 0.953 & 0.780 & 1.24 \\
227 & 52.9 & 1.000 & 25.2 & 0.977 & 0.886 & 1.13 \\
231 & 54.0 & 0.997 & 26.0 & 0.946 & 0.433 & 1.00 \\
232 & 53.1 & 1.000 & 26.7 & 0.956 & 0.759 & 2.44 \\
235 & 51.0 & 0.999 & 25.1 & 0.964 & 0.850 & 1.53 \\
242 & 53.4 & 0.995 & 25.5 & 0.922 & 0.323 & 1.15 \\
246 & 51.7 & 0.998 & 25.9 & 0.968 & 0.653 & 0.94 \\
\midrule
\multicolumn{7}{c}{Summary Statistics} \\
\midrule
Mean & 53.8 & 0.999 & 25.8 & 0.960 & 0.689 & 1.10 \\
Std & 1.2 & 0.001 & 0.6 & 0.014 & 0.165 & 0.41 \\
\bottomrule
\end{tabular}
\end{table}

\begin{table}[h!]
\centering
\caption{Detailed evaluation results for LIC-TCM model on DIV2K subset}
\label{tab:tcm_results_div2k}
\small
\begin{tabular}{@{}c|cc|cccc@{}}
\toprule
\multirow{2}{*}{\begin{tabular}[c]{@{}c@{}}Image\\(idx)\end{tabular}} & \multicolumn{2}{c|}{\textbf{Stealth} $\uparrow$} & \multicolumn{4}{c}{\textbf{Attack Success} $\downarrow$} \\
\cmidrule(lr){2-3} \cmidrule(lr){4-7}
& PSNR & VIF & PSNR & SSIM & VIF & BPP \\
\midrule
01 & 55.1 & 0.999 & 24.9 & 0.970 & 0.763 & 1.38 \\
16 & 54.8 & 1.000 & 25.4 & 0.964 & 0.829 & 1.40 \\
35 & 53.7 & 1.000 & 27.1 & 0.965 & 0.826 & 2.06 \\
49 & 53.3 & 1.000 & 25.3 & 0.975 & 0.833 & 1.69 \\
51 & 53.6 & 0.999 & 25.1 & 0.963 & 0.812 & 1.70 \\
53 & 54.9 & 1.000 & 26.3 & 0.961 & 0.821 & 2.19 \\
55 & 52.5 & 0.998 & 25.4 & 0.942 & 0.642 & 1.53 \\
90 & 54.9 & 1.000 & 26.1 & 0.971 & 0.866 & 0.99 \\
106 & 53.1 & 1.000 & 26.8 & 0.971 & 0.898 & 1.97 \\
110 & 55.0 & 0.999 & 26.3 & 0.962 & 0.671 & 0.91 \\
111 & 54.4 & 1.000 & 26.7 & 0.977 & 0.912 & 1.14 \\
115 & 51.8 & 0.999 & 26.1 & 0.958 & 0.680 & 1.78 \\
121 & 52.7 & 0.999 & 25.7 & 0.965 & 0.690 & 1.12 \\
123 & 51.3 & 0.999 & 25.6 & 0.955 & 0.808 & 1.44 \\
\bottomrule
\end{tabular}
\end{table}

\begin{table}[h!]
\centering
\caption{Detailed evaluation results for LIC-TCM model on DIV2K subset (\textbf{continued})}
\label{tab:tcm_results_div2k}
\small
\begin{tabular}{@{}c|cc|cccc@{}}
\toprule
\multirow{2}{*}{\begin{tabular}[c]{@{}c@{}}Image\\(idx)\end{tabular}} & \multicolumn{2}{c|}{\textbf{Stealth} $\uparrow$} & \multicolumn{4}{c}{\textbf{Attack Success} $\downarrow$} \\
\cmidrule(lr){2-3} \cmidrule(lr){4-7}
& PSNR & VIF & PSNR & SSIM & VIF & BPP \\
\midrule
127 & 54.9 & 0.999 & 27.1 & 0.962 & 0.732 & 1.08 \\
130 & 55.0 & 0.998 & 25.4 & 0.963 & 0.467 & 1.10 \\
138 & 52.8 & 0.992 & 25.1 & 0.965 & 0.216 & 1.08 \\
139 & 55.4 & 1.000 & 25.6 & 0.963 & 0.898 & 2.08 \\
141 & 54.3 & 1.000 & 25.1 & 0.970 & 0.860 & 1.81 \\
155 & 55.0 & 1.000 & 24.9 & 0.963 & 0.813 & 1.60 \\
201 & 52.0 & 1.000 & 26.4 & 0.955 & 0.770 & 2.05 \\
213 & 52.4 & 1.000 & 25.2 & 0.982 & 0.961 & 3.19 \\
229 & 55.0 & 0.999 & 26.1 & 0.960 & 0.634 & 1.24 \\
268 & 54.1 & 0.998 & 25.8 & 0.959 & 0.537 & 0.92 \\
269 & 53.2 & 0.999 & 25.5 & 0.947 & 0.708 & 1.13 \\
297 & 54.6 & 1.000 & 27.1 & 0.970 & 0.895 & 2.97 \\
299 & 54.9 & 1.000 & 25.6 & 0.974 & 0.894 & 1.93 \\
305 & 55.2 & 1.000 & 25.1 & 0.954 & 0.760 & 2.38 \\
324 & 53.1 & 0.999 & 26.5 & 0.954 & 0.804 & 2.10 \\
336 & 54.9 & 0.999 & 25.6 & 0.962 & 0.721 & 1.58 \\
349 & 54.9 & 1.000 & 25.2 & 0.958 & 0.868 & 2.30 \\
351 & 54.9 & 1.000 & 27.2 & 0.970 & 0.918 & 1.55 \\
363 & 54.2 & 1.000 & 27.0 & 0.968 & 0.888 & 2.52 \\
365 & 55.0 & 1.000 & 25.0 & 0.973 & 0.737 & 0.98 \\
374 & 55.0 & 1.000 & 27.0 & 0.962 & 0.773 & 1.70 \\
397 & 54.5 & 0.999 & 25.0 & 0.959 & 0.689 & 1.65 \\
404 & 51.9 & 0.999 & 25.6 & 0.942 & 0.809 & 1.66 \\
407 & 54.6 & 1.000 & 27.1 & 0.970 & 0.895 & 2.97 \\
412 & 53.6 & 1.000 & 26.3 & 0.963 & 0.916 & 2.02 \\
468 & 51.9 & 0.999 & 25.4 & 0.952 & 0.845 & 1.55 \\
502 & 52.2 & 1.000 & 26.2 & 0.970 & 0.863 & 2.25 \\
507 & 54.8 & 0.999 & 26.1 & 0.959 & 0.543 & 1.08 \\
524 & 54.8 & 1.000 & 25.2 & 0.975 & 0.834 & 0.70 \\
525 & 55.1 & 1.000 & 25.0 & 0.940 & 0.640 & 0.88 \\
533 & 55.0 & 1.000 & 26.9 & 0.970 & 0.915 & 1.58 \\
539 & 53.9 & 1.000 & 25.9 & 0.972 & 0.892 & 1.15 \\
578 & 54.8 & 1.000 & 26.6 & 0.960 & 0.690 & 1.09 \\
584 & 53.1 & 1.000 & 25.1 & 0.972 & 0.852 & 1.39 \\
592 & 54.9 & 1.000 & 26.6 & 0.951 & 0.801 & 2.02 \\
606 & 55.1 & 1.000 & 25.0 & 0.974 & 0.878 & 1.61 \\
618 & 53.0 & 0.999 & 25.5 & 0.971 & 0.755 & 0.96 \\
627 & 52.0 & 1.000 & 24.7 & 0.974 & 0.858 & 1.70 \\
632 & 55.1 & 0.998 & 24.9 & 0.963 & 0.534 & 1.23 \\
644 & 53.8 & 1.000 & 26.0 & 0.977 & 0.930 & 1.46 \\
657 & 55.0 & 1.000 & 25.8 & 0.956 & 0.903 & 2.31 \\
677 & 55.2 & 1.000 & 25.1 & 0.981 & 0.873 & 0.91 \\
681 & 54.2 & 1.000 & 25.1 & 0.980 & 0.913 & 1.22 \\
683 & 54.9 & 0.999 & 24.8 & 0.962 & 0.663 & 1.24 \\
727 & 52.5 & 0.998 & 25.5 & 0.958 & 0.612 & 1.09 \\
740 & 55.0 & 1.000 & 27.4 & 0.963 & 0.882 & 2.31 \\
749 & 54.3 & 0.999 & 24.9 & 0.975 & 0.793 & 0.66 \\
761 & 52.9 & 1.000 & 25.4 & 0.952 & 0.798 & 2.26 \\
784 & 51.9 & 0.999 & 26.9 & 0.950 & 0.816 & 1.29 \\
788 & 54.9 & 0.999 & 24.7 & 0.976 & 0.741 & 1.98 \\
797 & 53.3 & 1.000 & 25.8 & 0.966 & 0.860 & 2.38 \\
\midrule
\multicolumn{7}{c}{Summary Statistics} \\
\midrule
Mean & 54.0 & 0.999 & 25.8 & 0.965 & 0.789 & 1.67 \\
Std & 1.1 & 0.001 & 0.7 & 0.010 & 0.132 & 0.57 \\
\bottomrule
\end{tabular}
\end{table}

\end{document}